\documentclass[preprint,12pt]{elsarticle}
\usepackage{amsmath,amssymb,amsfonts}
\usepackage{algorithm}    
\usepackage{algpseudocode} 
\usepackage{graphicx}
\usepackage{textcomp}
\usepackage{longtable}
\usepackage{booktabs}
\usepackage{comment}
\usepackage{multirow}
\usepackage{graphicx}
\usepackage{subcaption}
\usepackage{dsfont}
\usepackage{url}
\usepackage{xcolor} 
\usepackage{tabularx}
\usepackage{float} 

\makeatletter
\def\ps@pprintTitle{%
   \let\@oddhead\@empty
   \let\@evenhead\@empty
   \let\@oddfoot\@empty
   \let\@evenfoot\@empty
}
\makeatother

\begin{document}

\begin{frontmatter}



\title{Mining-Gym: A Configurable RL Benchmarking Environment for Open-Pit Truck Dispatch Optimization}

\author[inst1]{Chayan Banerjee\corref{cor1}}
\author[inst1]{Kien Nguyen}
\author[inst1]{Clinton Fookes}

\affiliation[inst1]{organization={School of Electrical Engineering and Robotics, Queensland University of Technology},%
            addressline={2 George Street}, 
            city={Brisbane},
            postcode={4000}, 
            state={QLD},
            country={Australia}}

\cortext[cor1]{Corresponding author. Email: c.banerjee@qut.edu.au}


\begin{abstract}
\textcolor{black}{Optimizing the mining process—particularly truck dispatch scheduling—is a key driver of efficiency in open-pit operations. However, the dynamic and stochastic nature of these environments, marked by uncertainties such as equipment failures, truck maintenance, and variable haul cycle times, poses significant challenges for traditional optimization. While Reinforcement Learning (RL) shows strong potential for adaptive decision-making in mining logistics, practical deployment hinges on thorough evaluation in realistic, customizable simulation environments. Yet the absence of standardized benchmarking currently hampers fair algorithm comparison, reproducibility of results, and real-world applicability of RL-based solutions in open-pit mining.
To bridge this gap, we present Mining-Gym—a configurable, open-source benchmarking environment for training, testing, and evaluating RL algorithms in mining process optimization. Mining-Gym is built on Salabim-based Discrete Event Simulation (DES) and integrates with Gymnasium (formerly OpenAI Gym), enabling seamless adoption of state-of-the-art RL algorithms from Stable Baselines3. The framework captures mining-specific uncertainties such as equipment failures, queue congestion, and stochastic behavior through an event-driven decision-point architecture. Mining-Gym provides a graphical user interface (GUI) for intuitive parameter configuration, extensive data logging, and real-time visualization—supporting thorough performance analysis and standardized, reproducible evaluation of diverse RL strategies and heuristic baselines.
To validate benchmarking performance, we compare classical heuristics with RL-based scheduling across six scenarios ranging from normal operation to severe equipment failures. The results demonstrate that Mining-Gym is an effective, reproducible testbed for mining-logistics research—enabling fair evaluation of adaptive decision-making for real mines. The demonstrative experiments also show the strong performance potential of RL-trained schedulers, motivating further development in this direction.}

\end{abstract}




\end{frontmatter}

\section{Introduction}
\label{sec:introduction}
Mining process optimization aims to enhance efficiency and productivity by improving resource allocation, equipment scheduling, and material handling. However, these operations are inherently complex and uncertain, influenced by dynamic and stochastic factors such as equipment failures, fluctuating ore quality, and unpredictable environmental conditions. These characteristics introduce significant levels of imprecision and variability into decision-making, making it challenging for traditional optimization methods—such as linear programming and heuristics—to adapt effectively in real time. As a result, these conventional approaches often lead to inefficiencies and increased operational costs \cite{wang2024dynamic}.

Reinforcement Learning (RL), a branch of machine learning grounded in soft computing principles, offers a promising adaptive approach for handling such complexity. 
However, its adoption in mining remains limited, primarily due to the absence of standardized benchmarking environments that enable fair algorithmic comparison, reproducible experimentation, and systematic evaluation across diverse operational scenarios. Moreover, existing simulators are often proprietary \cite{komatsu_dispatch_fms_2024}, overly simplistic \cite{wang2023real}, or narrowly tailored to specific algorithms—factors that hinder fair comparisons and slow the transition from academic development to industrial application.

To bridge this gap, we present \textit{Mining-Gym}, a configurable benchmarking framework for truck dispatch scheduling optimization, designed specifically to support standardized evaluation of RL and classical algorithms in mining contexts. Unlike prior tools, Mining-Gym provides a comprehensive and open-source platform that supports the development and evaluation of RL algorithms by incorporating:

\begin{itemize}
\item A high-fidelity discrete-event simulation (DES) capturing real-world operational complexities and stochastic behaviors,
\item Seamless integration with OpenAI Gym to ensure compatibility with modern RL algorithms,
\item Extensive customization options to model diverse and realistic mining scenarios,
\item A comprehensive system for logging, visualization, and performance tracking, and
\item Open-source~\footnote{Source code available at https://github.com/CBCodeHub/Mining-Gym} availability to promote transparency, collaboration, and reproducibility in research.
\end{itemize}

By enabling standardized benchmarking and fostering robust comparisons across algorithms and heuristics, Mining-Gym supports the practical application of RL in mining logistics. It serves as a crucial bridge between theoretical research in soft computing and its deployment in real-world industrial systems. The framework's primary contribution is providing researchers with a reliable testbed for rigorous algorithmic evaluation. \textcolor{black}{As a proof-of-concept demonstration, we present comparative results between classical heuristics (random, fixed) and RL-based scheduling (PPO) across six operational scenarios, validating Mining-Gym's effectiveness as a benchmarking platform.}

\textcolor{black}{The remainder of this paper is organized as follows: Section \ref{Sec:II} reviews related work in mining simulation and reinforcement learning, emphasizing the limitations of current benchmarking approaches. Section \ref{Sec:III} describes the architecture of Mining-Gym, which integrates discrete-event simulation with RL to facilitate standardized evaluation. Section \ref{Sec:IV} presents the implementation details, including the configuration system, visualization tools, and logging mechanisms that ensure reproducible experiments. Section \ref{Sec:V} outlines the experimental setup and provides a demonstrative comparison between classical heuristics and RL-based scheduling to showcase the framework's benchmarking capabilities. Section \ref{sec:VI} discusses the results of these experiments. Finally, Section \ref{sec:VII} concludes the paper and highlights directions for future work.}

\section{Background and Related Work}\label{Sec:II}
\subsection{Background}
\textcolor{black}{
\subsubsection{Truck dispatch scheduling optimization} Open-pit mining extracts ore through a coordinated process: hydraulic shovels 
excavate ore and waste at the pit face, trucks transport material to processing 
facilities, crushers process the ore, and dumping sites stockpile waste. The 
\textbf{dispatch problem} focuses on optimizing truck-to-shovel assignments  to minimize idle time while respecting equipment constraints. Key operational features in the mining setting are: 
\begin{itemize}
    \item Multi-destination routing: Ore is routed to crushers and waste to dumping sites; the dispatcher must balance this allocation.
    \item Equipment capacity constraints: Shovels and crushers have finite capacities, leading to potential queues and bottlenecks.
    \item Stochastic disruptions: Equipment failures and variable cycle times require adaptive, real-time decision-making.
    \item Cost drivers: Truck idle time dominates operational costs; minimizing queue congestion is therefore economically critical.
\end{itemize}
Dispatching decisions significantly influence operational efficiency, as a large 
portion of mining costs are linked to truck–shovel activities. Truck dispatching tasks 
often employ mathematical programming approaches to reduce equipment waiting times 
and optimize production \cite{ta2005stochastic}.}

Heuristic methods are widely used for their simplicity and flexibility, especially in dynamic mining environments where traditional optimization struggles with real-time disruptions like equipment failures or changing conditions \cite{fang2023micro}. Common strategies include assigning trucks to the nearest shovel, prioritizing by equipment capacity or material demand, and using historical data trends. However, they lack guarantees of global optimality and may underperform in complex or unfamiliar scenarios.
In contrast, static optimization approaches are often applied to truck dispatching tasks to minimize equipment idle time and enhance production efficiency \cite{ta2005stochastic}. However, these traditional methods usually require full re-optimization when faced with complex changes—like equipment breakdowns—and are often not robust enough to manage the inherent uncertainty of mining operations.

Given the limitations of static optimization methods in handling dynamic and uncertain mining environments, there is a growing research emphasis on developing adaptive and flexible approaches that can respond effectively to real-time operational disruptions.

\subsubsection{DES-RL integration in Mining Operations}
To effectively apply RL in mining logistics, it must be integrated with robust simulation frameworks. DES has been widely used in industrial and mining applications for optimizing processes like truck scheduling. By modeling stochastic interactions between equipment and processors, DES captures variability and complexity, often replacing intricate mathematical models with probabilistic parameters \cite{law2007simulation}. Applications include supply chain evaluation \cite{bodon2018combining} and ensuring adherence to production schedules \cite{manriquez2020simulation}, demonstrating DES’s role in handling uncertainty and improving operational efficiency.

Recent research has explored combining RL with DES to simulate complex industrial environments. While standard RL environments like OpenAI Gym provide useful benchmarks, they often lack the realism required for industrial applications \cite{hubbs2020or}. Industrial settings demand detailed, stochastic modeling, similar to DES, to account for dynamic conditions and resource constraints. Integrating RL’s trial-and-error learning with DES has gained traction for modeling real-world stochastic systems. For example, \cite{zielinski2021flexible} transformed DES-based SCT (supervisory control theory) controllers into RL environments, enhancing decision-making in automotive plant control.

This RL-DES integration has been applied to scheduling and optimization across industries. \cite{lang2020integration} employed Deep Q-Networks (DQN) for flexible job shop problems, outperforming traditional metaheuristics, while \cite{lang2021modeling} used DES and OpenAI Gym to create RL-compatible production scheduling environments, simplifying RL algorithm deployment.
In mining, RL has been explored for short-term planning, truck dispatching, and scheduling. \cite{zhang2023vehicle} introduced a curriculum-driven RL method for vehicle dispatching to address sparse rewards, while \cite{matsui2023real} developed a real-time RL-based dispatching system for autonomous trucks. Additionally, \cite{chiarot2024improving} applied Q-learning to optimize material supply during operational delays. These studies highlight the potential of RL-DES integration in enhancing decision-making and efficiency in mining operations.


\subsubsection{RL and the importance of simulator}

RL  excels in sequential decision-making, delivering state-of-the-art performance across various domains, including robotics, locomotion control, autonomous driving, and multi-agent systems \cite{sutton2018reinforcement}.
%
The mining truck dispatching problem can be framed as a sequential decision-making task, where truck assignments must adapt to evolving conditions. RL provides a suitable framework for optimizing these dispatching strategies 

In mining, RL enables adaptive decision-making by learning optimal dispatching policies through trial-and-error interactions. This approach accommodates dynamic changes in configurations, equipment failures, fluctuating ore quality, and weather, without frequent re-optimization. RL allows continuous refinement of dispatching policies to maximize efficiency and resilience in complex, uncertain environments.
However, RL’s reliance on environmental interaction presents challenges in real mining due to safety, cost, and timeline concerns. Unlike controlled simulations 
real-time training is impractical as RL requires extensive exploration. Despite advancements in sample efficiency \cite{banerjee2023enhancing,banerjee2024improved}, mining operations still require hundreds to thousands of episodes, hindering real-world deployment.

Simulators are preferred for safe, cost-effective RL training. They enable rapid learning in diverse virtual scenarios, crucial for robust policy development. Common RL simulators like OpenAI Gym \cite{brockman2016openai}, 
although simplified, form the basis for algorithm development.
Simulators are vital in mining optimization due to industry complexity and variability. High-fidelity simulations improve accuracy, ensuring learned strategies transfer effectively to real-world operations, enhancing efficiency and safety, making RL a viable tool for mining logistics.

\subsubsection{RL Applications in Truck Dispatch Optimization}
A limited number of studies have applied RL to truck dispatch optimization, but these are typically tailored to specific RL algorithms, limiting their adaptability. Adapting them to work with different RL methods would require extensive understanding and potential modifications.
Huo et al. \cite{huo2023reinforcement} apply Q-learning to optimize dispatching in haulage operations, reducing greenhouse gas emissions while maintaining production. Matsui et al. \cite{matsui2023real} develop a real-time dispatching algorithm using deep RL for autonomous haulage trucks, improving transportation efficiency and fuel consumption. De et al. \cite{de2023integrating} extend RL to short-term production planning, integrating actor-critic agents for equipment allocation and production scheduling. Chiarot et al. \cite{chiarot2024improving} apply Q-learning-based deep RL to reduce delays during shifts and breaks, improving material supply to crushers.

\begin{table*}[t]
\centering
\scalebox{0.57}{
\begin{tabular}{|l|l|l|l|l|l|l|l|l|}
\hline
Ref.                  & Year        & Uncertainties         & Simulator arch.            & RL adaptability         & RT Visualization         & Customizable        & Platform used         & OS*         \\ \hline
\cite{zhang2022real}         & 2022      & Extensive & DES          &  N.A.                &    $\times$             &    $\checkmark$        & Simpy      &    $\times$             \\ \hline
\cite{zhang2022determination} & 2022      & Extensive & DES          &      N.A.           &  Fair       &    $\checkmark$        & Flexsim    &    $\times$             \\ \hline
\cite{dendle2022efficient}   & 2022      & Extensive & DES          &    N.A.             &     $\times$            &    $\checkmark$        & -          &  $\times$               \\ \hline
\cite{zhang2023vehicle}      & 2023      & Fair      & DES          & Custom     &     $\times$           &     $\checkmark$       & Simpy      &      $\times$           \\ \hline
\cite{huo2023reinforcement}  & 2023      & Fair      & Rule based   & Custom     &    $\times$             &    $\times$            & OpenAI     &    $\times$             \\ \hline
\cite{matsui2023real}        & 2023      & Fair      & DES          & Custom     &    $\times$             &     $\checkmark$       & Python     &    $\times$             \\ \hline
\cite{wang2023real}          & 2023      & Fair      & Rule based   &     N.A.              &     $\times$            &     $\times$            & MATLAB     &   $\times$              \\ \hline
\cite{de2023integrating}     & 2023      & Extensive & Rule based   & Custom     &    $\times$             &     $\times$            & -          &       $\times$          \\ \hline
\cite{moradi2024nested}      & 2024      & Fair      & DES          &      N.A.             & Fair       &      $\checkmark$      & Arena      &  $\times$               \\ \hline
\cite{meng2024openmines}     & 2024      & Extensive & DES          &     N.A.            & Extensive  &   $\checkmark$         & Simpy      &     $\checkmark$           \\ \hline
\cite{chiarot2024improving}  & 2024      & Fair      & DES          & Custom     & Fair       &   $\checkmark$         & DISPATCH  &  $\times$   \\
\hline
\textbf{Ours} & \textbf{2025}      & \textbf{Extensive}    & \textbf{DES}        & \textbf{Adaptable}  & \textbf{Extensive}  &   $\checkmark$   & \textbf{Salabim, Python}  &  $\checkmark$  \\ \hline
\end{tabular}
}
\caption{
Comparative analysis of mining simulators for truck dispatching based on key features. The table evaluates simulators across several dimensions: \textbf{Uncertainties}, ranging from \textit{Extensive} (comprehensive modeling of equipment failures, maintenance, variable haul times) to \textit{Fair} (modeling of limited uncertainties); \textbf{Simulator Architecture}, comparing \textit{(DES)} and \textit{Simpler Rule-based Logic}; \textbf{RL Adaptability}, which indicates whether the simulator is \textit{Adaptable} (native integration), \textit{Custom} (requires adaptation), or \textit{N.A.} (not designed for RL); \textbf{Real-time Visualization}, measuring the extent of \textit{Extensive} (Elaborate animated visuals with KPI dashboards), \textit{Fair} (basic visual representation, or \textit{None}. *\textit{OS }stands for Open Source/ public availability of code or simulator for use.}
\label{tab:simulation_studies}
\end{table*}

\subsubsection{Limitations: Existing simulators}
Refer to Table~\ref{tab:simulation_studies} for a comparison of conventional simulators based on crucial features. A notable absence of real-time visualization is observed in several studies, including \cite{zhang2022real,dendle2022efficient,zhang2023vehicle,huo2023reinforcement,matsui2023real,wang2023real,de2023integrating,chiarot2024improving}, which can hinder the ability to monitor and manage mining operations effectively. Customizability is another missing feature in \cite{huo2023reinforcement}, \cite{wang2023real}, and \cite{de2023integrating}, limiting the flexibility of these simulators for adapting to different mining scenarios. Furthermore, the scarcity of open-source solutions, with only \cite{meng2024openmines} providing such a framework, restricts broader adoption and collaborative improvement. A number of works, such as \cite{huo2023reinforcement}, \cite{wang2023real}, and \cite{de2023integrating}, utilize rule-based architectures that may not be as adaptable or robust as more advanced simulation methods. Additionally, the proprietary nature of many conventional simulators limits research replication and comparison. Many also fail to adequately account for random events affecting key mining components like trucks, shovels, or crushers, as seen in \cite{huo2023reinforcement}. Finally, many simulators are not designed for RL settings or are tailored to specific algorithms, limiting their versatility \cite{matsui2023real}.

Addressing these limitations is crucial for effectively training and testing RL algorithms to obtain optimal policies, for surface mining process optimization. Proper real-time visualization and comprehensive data logging are essential for ensuring the repeatability and comparability of experiments. Accurate simulation of real-world complexities and uncertainties, along with seamless integration with widely accepted formats like OpenAI Gym, for ease of use and training of off-the-shelf and custom algorithms. 

\begin{figure*}[h!]
    \centering
    \begin{subfigure}[t]{\linewidth}
        \centering
        \includegraphics[width=0.8\linewidth]{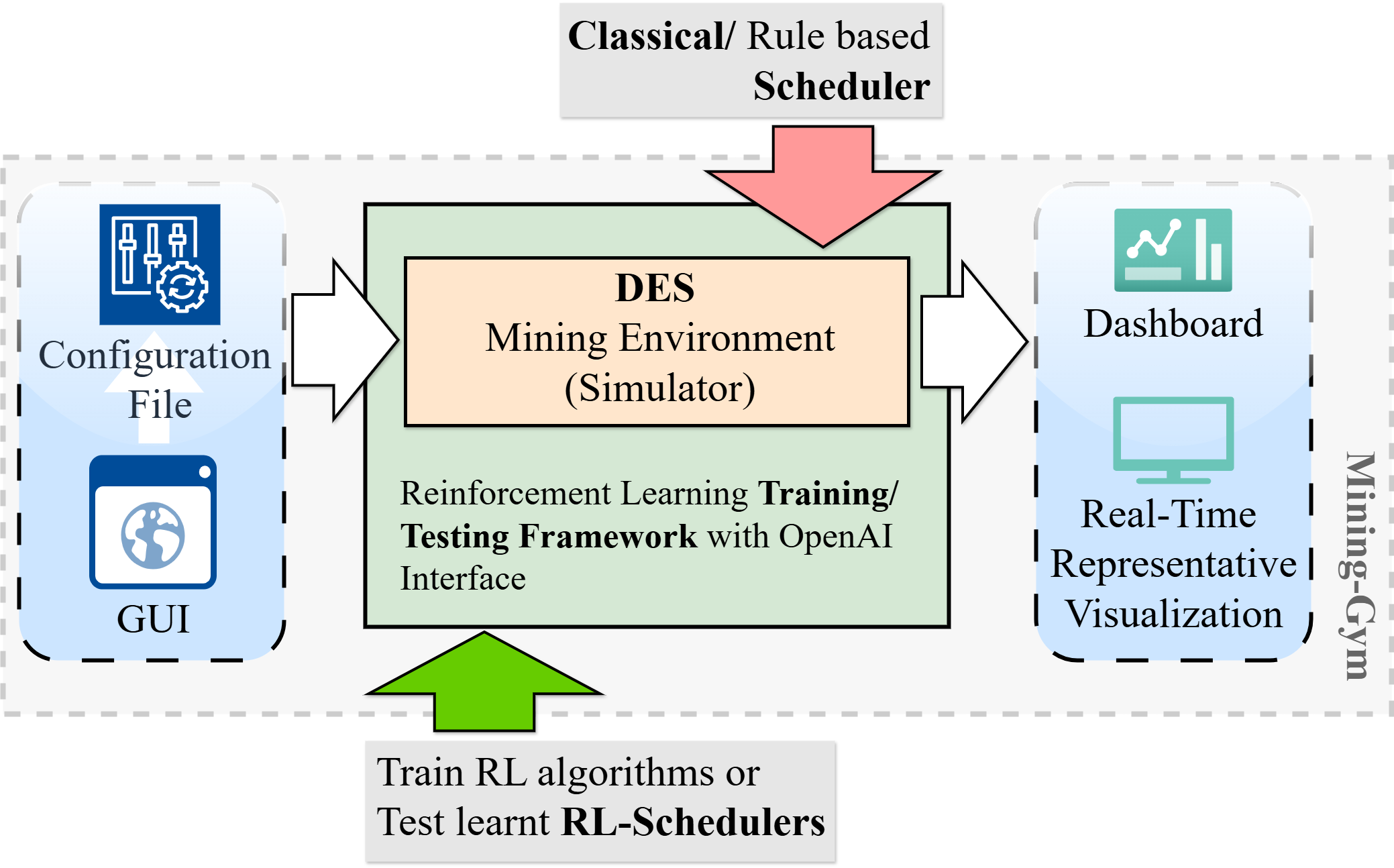}
        \caption{}
        \label{fig:interface_top}
    \end{subfigure}
    
    \vspace{0.5em} 

    \begin{subfigure}[t]{\linewidth}
        \centering
        \includegraphics[width=0.84\linewidth]{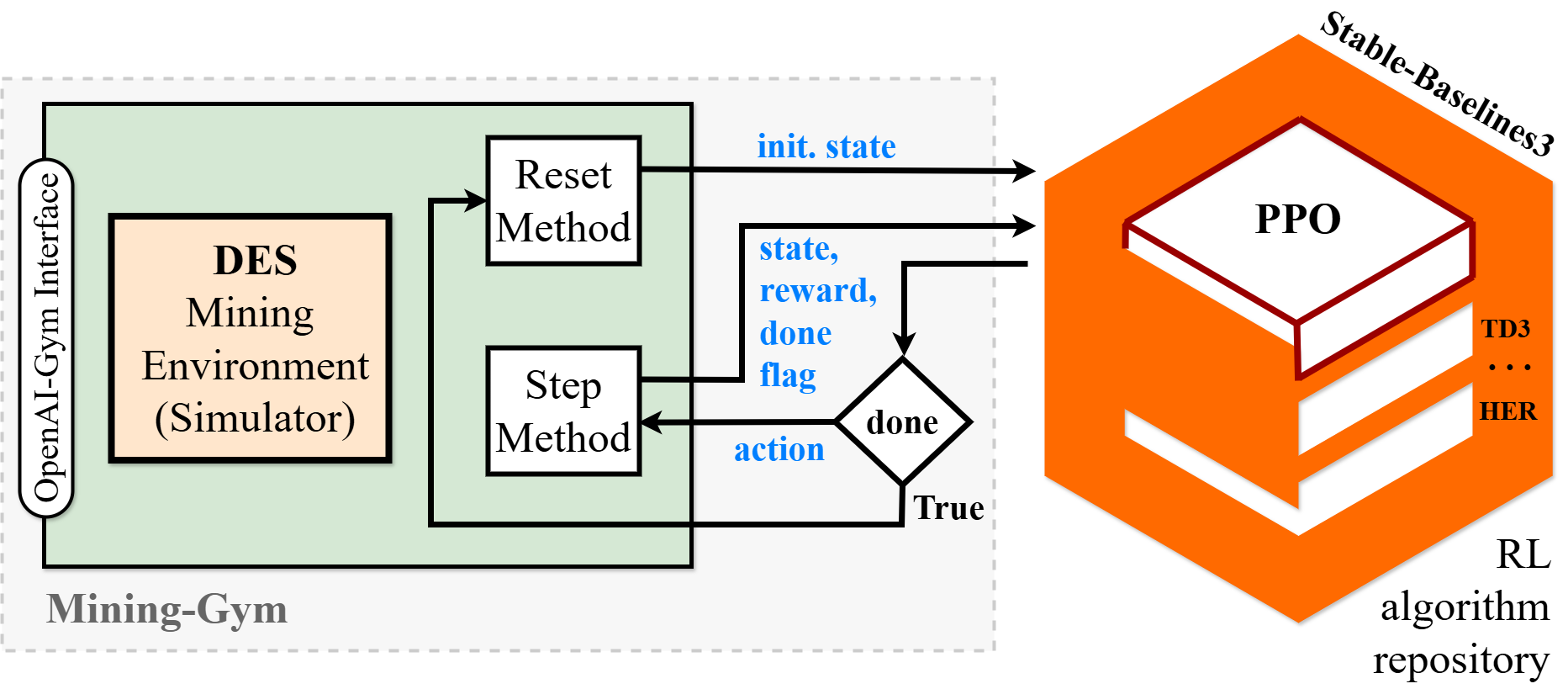}
        \caption{}
        \label{fig:interface_bottom}
    \end{subfigure}

    \caption{(a) Comprehensive system architecture: Displays all key components of the Mining-Gym, including the graphical user interface (GUI), the generated configuration file used to initialize the environment, and the dashboard with real-time visualizations.
    (b) OpenAI Gym-compatible RL interface: Illustrates how the Mining-Gym integrates with OpenAI Gym by adapting simulation signals into the standard reset and step methods. This compatibility allows seamless integration with popular RL libraries, such as Stable-Baselines3 \cite{stable-baselines3}, enabling easy training and testing of RL models.}
    \label{fig:levels}
\end{figure*}

\section{Mining-Gym: System Architecture and Design}\label{Sec:III}
Mining-Gym’s architecture integrates two complementary modeling approaches to tackle the truck dispatch scheduling challenge. It captures the dynamic, stochastic Load-Haul-Dump (crusher or dumping site)-Return-Queue (LHDRQ) cycle via a high-fidelity Discrete Event Simulation (DES) model, while the dispatch scheduling is formulated as a Markov Decision Process (MDP), enabling Reinforcement Learning (RL) for adaptive optimization. This dual-modeling framework represents both the operational complexity of mining and the sequential decision-making essential for intelligent dispatching. The following sections detail Mining-Gym’s architecture, covering the system overview, DES implementation, and MDP formulation for RL-based optimization.

\subsection{Mining-Gym overview}
Mining-Gym is a mining gym simulator and benchmarking tool designed for RL  applications in mining operations. \textcolor{black}{It features a GUI-based interface where users input crucial parameter values that define the state of the mining site, equipment, and other relevant factors. These parameters are translated into a human-readable text file, where they are divided into fixed parameters (sampled once for the entire run, defining the infrastructure) and operational parameters (stochastically sampled per shift/episode, modeling daily variability). This structure allows for easy adjustments directly within the simulation environment, while separate scenario files are now used to manage failure conditions for clearer experimentation.} Users can choose to either train a scheduler policy or run an existing one, which can be either classical or RL-based. The simulator includes an additional interface that supports the training and testing of modern RL algorithms, enabling advanced simulation capabilities. Furthermore, the tool provides a real-time dashboard displaying key performance indicators (KPIs) and a dynamic, visual representation of the mining site’s operations, offering users valuable insights into the system’s performance and the effects of various parameters on mining activities (see Fig.~\ref{fig:levels}).

We present a DES environment for mining processes, integrated with OpenAI Gym, (currently Gymnasium) —for RL  applications. Gymnasium is a widely used toolkit that standardizes interactions across diverse environments, from control tasks to robotics and video games. It enhances the original Gym with improved modularity, better support, and expanded features, enabling efficient testing, comparison, and debugging of RL algorithms. Our wrapper maps DES inputs/outputs to Gym’s reset and step methods, ensuring compatibility with RL frameworks like Stable Baselines. 
Stable Baselines provides well-tested implementations of RL algorithms such as PPO, A2C, and DQN, simplifying integration with Gym environments. The done flag signals episode termination, while info offers additional simulation insights. Together, Gymnasium and Stable Baselines create a powerful ecosystem for RL research, facilitating reproducible experimentation and benchmarking in complex domains like mining simulations.

\subsection{Mining Process Modeling: DES Overview}
For DES modeling, we have used a comprehensive python package named Salabim 
\footnote{Salabim is an open-source DES and animation library in Python. Available at \url{https://www.salabim.org/}} which offers process interaction methods, queue handling, resources, statistical sampling, and real-time 2D/3D animation capabilities.\newline
\textit{Components} are fundamental building blocks that define the dynamic behavior of entities within the simulation environment. By defining entities as components, Salabim can simulate complex interactions, resource contention, and event-driven behaviors essential for realistic modeling. Trucks (lower priority) and Breakdown Events (higher priority) are modeled as components in our work.
\textit{Resource} is a fundamental component used to model and manage entities that are shared among components within a simulation. Resources represent facilities, equipment, or services that components (such as trucks, shovels, or processes) compete for or utilize during their activities. Specifically, in Mining-Gym, Shovels, Dumps and Crushers are modeled as resources.

\textcolor{black}{
In the Salabim framework, the Truck component follows a Load–Haul– Dump–Return–Query (LHDRQ) cycle with stochastic parameters representing operational variability.\newline
In the \textbf{Loading phase}, a truck requests an available shovel as advised by the \textbf{Dispatcher} (classical or RL based) and waits for its turn. Once granted access, it undergoes loading, with material transfer time sampled from a configurable probability distribution (see Table~\ref{tab:Distribution} for example parameterizations).
The truck then enters the \textbf{Haul phase}, traveling a predefined trajectory to the dump (crusher or dumping site), represented by holding for travel time. Travel time depends on load state: loaded trucks travel at reduced speed due to fuel penalties, while empty returns occur at higher speed. Both speeds are sampled from configurable distributions to reflect operational variability.
At the dump, the truck enters the \textbf{Dump phase}, requesting access to an available crusher or dumping site. After access, it performs dumping, with unloading time sampled from distributions that differ between crushers and dumping sites to reflect their operational characteristics. The \textbf{Return phase} follows, where the truck travels back to the shovel, completing one haul cycle.
\textbf{Stochastic disruptions} can interrupt any phase: equipment failures occur according to Mean Time Between Failure (MTBF) distributions (typically modeled as Poisson processes), triggering the preemption handler that suspends operations until repairs complete after a Mean Time To Repair (MTTR) sampled from appropriate distributions. The \textbf{Query phase} then begins, where the truck seeks scheduling from the dispatcher. This event-driven approach ensures dispatch decisions account for truck capacity constraints, variable cycle times, and disruption-induced uncertainty.
All distributional parameters are fully configurable through the Mining-Gym interface, allowing users to calibrate the simulation to specific mine characteristics or test algorithmic robustness across diverse operational settings.
}

At the dump, the truck enters the \textbf{Dump phase}, requesting access to an available crusher or dumping site. After access, it performs dumping, mirroring the time needed to unload ore or waste. The return phase follows, where the truck travels back to the shovel, completing one haul cycle. The query phase then begins, where the truck seeks scheduling or resource allocation from the dispatcher.
The \textbf{Dispatcher agent} is the core scheduling strategist, optimizing mining operations by balancing resource utilization and minimizing wait times. It employs basic strategies like fixed schedules or nearest-first, as well as advanced approaches such as RL-trained neural network policies.
A truck may request the following resource allocations: 1) Shovel, 2) Crusher, 3) Dumping site, and 4) Route (not currently considered). Mining-Gym’s dispatcher modules handle these requests (except routing) and supports baseline (e.g. Random) aswell as learned strategies (RL based). 

\begin{figure*}[t!]
    \centering
    \includegraphics[width=0.9\linewidth]{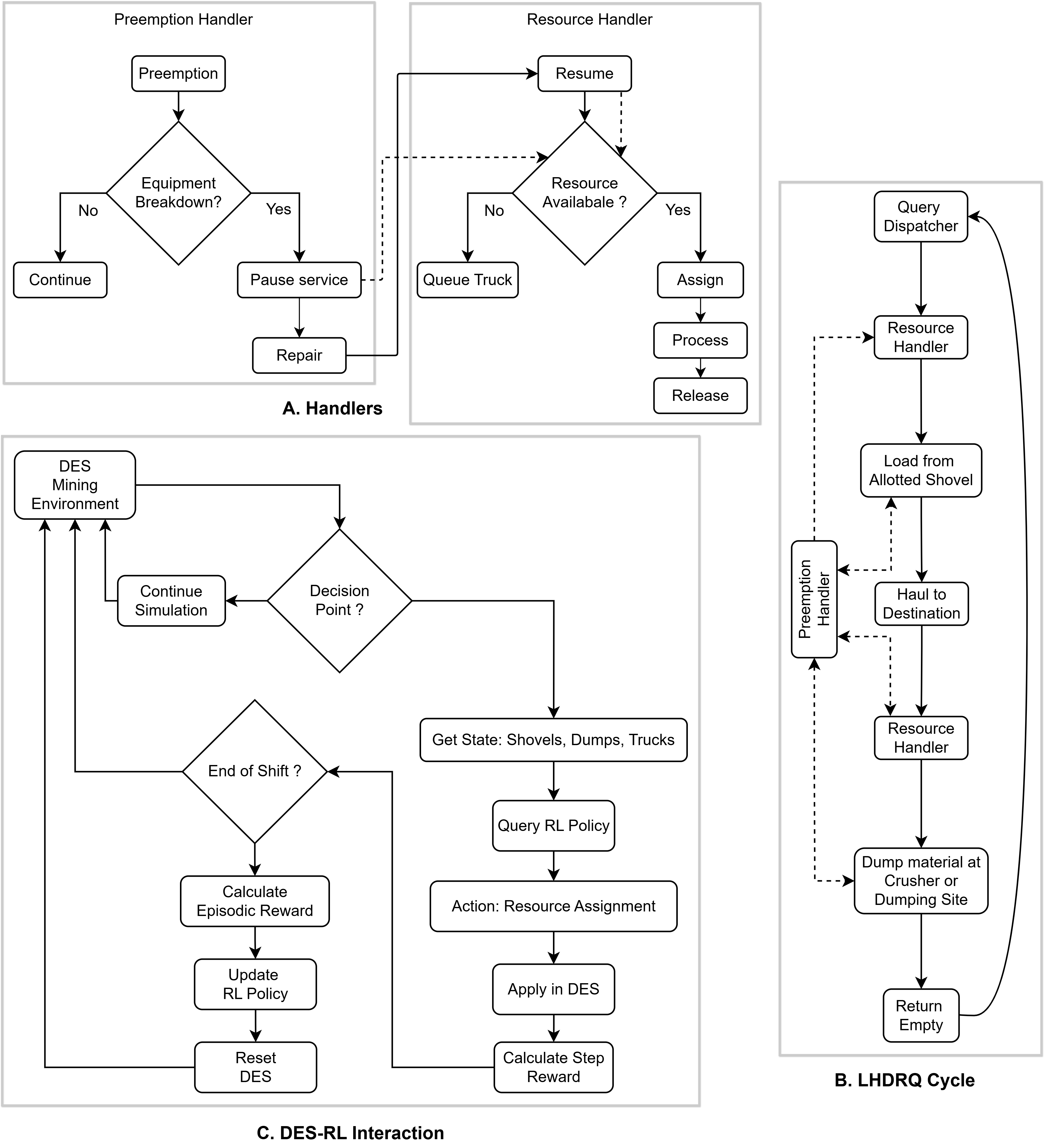}
    \caption{(A) Simplified mine-site simulation logic of Mining-Gym, showing three key components: (1) Resource Handler managing resource availability and assignments, (2) Preemption Handler detecting breakdowns and managing repair processes (B) Load-Haul-Dump-Return-Query (LHDRQ) cycle illustrating the truck's journey through the mining process, which begins with querying the dispatcher for assignments, followed by loading material, hauling to the destination, dumping, and returning empty. Breakdown events, managed by the Preemption Handler, can interrupt operations at any stage. (C) DES-RL interaction flow illustrating how the RL policy integrates with the DES. At decision points, the environment state is processed by the RL policy to determine resource assignments. Immediate or step rewards guide learning during simulation, while the episodic reward at shift (or episode) end updates the policy before environment reset. }
    \label{fig:Haul_logic}
\end{figure*}

\begin{algorithm}[H]
\caption{MineGym: Event-Driven DES-RL Synchronization}
\label{alg:des_gym}
\small 
\begin{algorithmic}[1]
\Require Simulation shift duration $T_{\text{shift}}$
\Ensure Trajectories $\{(s_t, a_t, r_t)\}_{t=0}^{T_{\text{shift}}}$

\State Initialize DES $\mathcal{S}_{\text{DES}} \gets \texttt{Salabim.Environment()}$
\State Initialize RL Agent $\mathcal{A}_{\text{RL}} \gets \texttt{Gym.Agent()}$
\State $t_{\text{current}} \gets 0$

\While{$t_{\text{current}} < T_{\text{shift}}$}
\State // \textbf{Decision Point Trigger (DES Pauses)}
\State $\texttt{event} \gets \mathcal{S}_{\text{DES}}.\texttt{run\_until\_next\_decision\_event()}$
\State $\tau \gets \mathcal{S}_{\text{DES}}.\texttt{event\_time\_advance}$
\State $d \gets \texttt{event.\text{triggering\_entity}}$ (e.g., Truck requesting assignment)
\State \Comment{The DES execution is instantaneously paused at $t = \mathcal{S}_{\text{DES}}.\texttt{now()}$}

\If{$d.\texttt{is\_initial\_state}$}
\State $a_t \gets \pi_{\text{classic}}(\text{shovel states})$ \Comment{Use default policy for initialization}
\Else
\State // \textbf{RL Observation}
\State $s_t \gets \mathcal{S}_{\text{DES}}.\texttt{get\_instantaneous\_state}(d)$ \Comment{Snapshot of system state at exact moment of pause}
\State // \textbf{Synchronous Communication (Gym Step)}
\State $\left(a_t, r_t\right) \gets \mathcal{A}_{\text{RL}}.\texttt{step\_and\_block}(s_t, \tau)$ \Comment{Agent computes action $a_t$, DES thread blocks until action received}
\State \Comment{Reward $r_t$ is computed by the Agent, often using $\tau$}
\EndIf

\State // \textbf{DES Resumes}
\State $d.\texttt{assign\_resource}(a_t)$ \Comment{Apply the chosen action to the truck entity}
\State $\mathcal{S}_{\text{DES}}.\texttt{resume\_and\_advance\_time}(\tau)$
\State $t_{\text{current}} \gets t_{\text{current}} + \tau$
\EndWhile

\end{algorithmic}
\end{algorithm}


\textcolor{black}{Additional features: The Mining-Gym workflow integrates DES modeling 
with RL-based decision-making through several key mechanisms illustrated in Fig.~\ref{fig:Haul_logic} and Algorithm~\ref{alg:des_gym}.\newline
Core System Architecture (Fig.~\ref{fig:Haul_logic}(A)): The simulation operates through 
two critical handlers: (1) the Resource Handler manages resource availability, 
queuing, and assignments for shovels, crushers, and dumps; (2) the Preemption 
Handler detects equipment breakdowns and coordinates repair processes, ensuring 
realistic disruption modeling across all equipment types.\newline
Truck Operation Cycle (Fig.~\ref{fig:Haul_logic}(B)): Trucks follow the Load-Haul-Dump-Return-
Query (LHDRQ) cycle, where each phase—from loading at shovels through hauling 
to dumps/crushers and returning—can be interrupted by breakdown events managed 
by the Preemption Handler. This event-driven architecture captures operational 
uncertainties inherent in mining logistics.\newline
DES-RL Synchronization (Fig.~\ref{fig:Haul_logic}(A) \& Algorithm~\ref{alg:des_gym}): At decision points (when trucks query for assignments), the DES pauses and transfers the current system 
state to the RL policy (Algorithm~\ref{alg:des_gym}, lines 6-14). The policy computes resource 
assignments while the DES thread blocks (line 16), ensuring synchronous 
communication. Immediate rewards guide step-wise learning, while episodic 
rewards at shift end enable policy updates before environment reset (lines 
17, 22). The
complete workflow integrates several advanced features designed to enhance
realism and flexibility, which are presented as follows.
}
\begin{itemize}
    \item \textit{Choice between Dump Types:} The stripping ratio of waste to ore is represented by the parameter $\epsilon$, which controls the probability that a truck carrying ore is directed to the crusher rather than the dumping site. Specifically, with probability $\epsilon$, the truck is directed to the crusher, and with probability $1 - \epsilon$, it is sent to the dumping site. This parameter can be adjusted to balance waste and ore management, adapting to varying operational conditions.
    
    \item \textit{Scalable Configurations:} The simulation supports adjustable configurations for Trucks, Shovels, Crushers, and Dumping sites, allowing for scalable operational simulations that can be tailored to different mining scenarios.
    
    \item \textit{Uncertainty Modeling:} To reflect the stochastic nature of real mining operations, the simulation incorporates sampling from various probability distributions. These distributions, along with their parameters, are configurable, enabling flexible modeling of operational uncertainty.
    
    \item \textit{Dispatcher Strategies:} Non-learnable baseline dispatching strategies, including the default random policy, are incorporated. These baselines enable systematic evaluation of different dispatching approaches and provide a reference for comparing the performance of RL-based policies
\end{itemize}

For detailed modeling assumptions underlying our DES implementation, please refer to Appendix A.

\subsection{Modeling the Dynamic Dispatching Problem as MDP}
\subsubsection{Typical RL}
In a typical RL  setting, the agent interacts with the environment at every time step \( t \). The environment is modeled as a Markov Decision Process (MDP) defined by the tuple \( (S, A, P, R) \), where \( S \) is the set of states, \( A \) is the set of actions, \( P(s'|s, a) \) is the state transition probability, and \( R(s, a) \) is the reward function. At each time step, the agent selects an action \( a_t \) based on the current state \( s_t \), receives a reward \( r_{t+1} \), and transitions to a new state \( s_{t+1} \). The goal is to learn a policy \( \pi(a|s) \) that maximizes the expected cumulative reward \( \mathbb{E}[\sum_{t=0}^\infty \gamma^t r_t] \), where \( \gamma \) is the discount factor.

\subsubsection{Decision-point based or Event-driven RL}
In mining truck scheduling, an event-driven reinforcement learning (RL) approach is adopted, where the agent interacts with the environment only at discrete decision points \(d\), rather than at every time step. A decision point represents a specific instance within the operational environment where the agent must choose an action based on the observed state \(s\), which includes truck and job statuses. The reward \(r\) at each decision point reflects performance metrics such as reduced waiting times and increased throughput, while the action \(a\) involves dispatching trucks to different tasks. The framework aims to optimize the expected cumulative reward, given by \( \mathbb{E}[\sum_{d=0}^\infty \gamma^d r_d] \), with agent-environment interactions constrained to these discrete decision points to align with real-world operational constraints.
This approach has been widely used in complex environments where continuous interaction is impractical or unnecessary, e.g. in vehicle routing \cite{hildebrandt2021action} 
and supply chain management \cite{hammler2023fully}.

\subsubsection{Defining Agent-Environment interaction and the MDP}
Formulating truck dispatch scheduling in open-pit mining as a Markov Decision Process (MDP) involves defining the components of MDP: states, actions, rewards, and transitions. 
The Mining-Gym framework considers $TR$ trucks (represented as $\tau$) interacting with the DES based mining environment. At every decision point $d \in D$, after dumping the material into the appropriate location (crusher or dumping site), a new resource assignment is requested by a truck $\tau_i\; \text{where}\; i \in N$. The dispatcher agent $\mathcal{D}$ observes the current state $S_d \in \mathcal{S}$, where $S_d$ represents the current status of the mining complex's performance at decision point $d$, and takes an action $A_d^i \in \mathcal{A}$, determining the next shovel to which the truck $i$ will be assigned. \newline

Below, we define the various components of the MDP used in this work:
\begin{itemize}
    \item [1)] \textit{States:} The state represents the current status of the mining operation. The state of the system at a given time must encode all the features needed for the agent to learn a relationship with the desired objective to be maximized. For this task, the state of the system is encoded as a vector with the following components: 
    \begin{align}
        s^{d} = [SA_{d}, TA_{d} ],
    \end{align}
    where $SA_d$ represents shovel-related attributes as an encoded vector at the current decision point $d$. It includes \textit{Shovel ID}, encoded in 3-bit chunks, \textit{Queue Length}, representing the number of trucks waiting per shovel, and \textbf{Shovel Status}, a binary indicator of shovel availability (online/offline). Similarly, $TA_d$ captures truck-related attributes, including \textit{TruckID}, encoded in 5-bit chunks, \textit{Trips complete}, tracking the number of trips per truck, and \textit{Trip Status}, a multi-bit representation of the truck's current state, such as loading, transit, or maintenance (see Table~\ref{tab:state_attrib_code}).

\begin{table}[h!]
    \centering
    \renewcommand{\arraystretch}{1.3}
    \resizebox{\textwidth}{!}{%
    \begin{tabular}{|l|l|p{3.5cm}|}
        \hline
        \textbf{Category} & \textbf{Attribute / Status} & \textbf{Description} \\ 
        \hline \hline
        \multirow{3}{*}{Shovels $(SA_{d})$}  
            & Shovel ID & 4-bit binary (supports up to 16 shovels) \\ \cline{2-3}
            & Queue Length & Normalized float [0, 1] (bounded by $2 \times N_{trucks}$) \\ \cline{2-3}
            & Shovel Status & Binary (1=operational, 0=breakdown) \\ 
        \hline
        \multirow{3}{*}{\parbox{2.5cm}{Active Truck\\$(TA_{d,active})$}}  
            & Truck ID & 6-bit binary (supports up to 64 trucks) \\ \cline{2-3}
            & Trips Complete & Normalized float (scaled by 500) \\ \cline{2-3}
            & Truck Status & 3-bit binary \\ 
        \hline
        \multirow{3}{*}{\parbox{2.5cm}{Fleet Context\\$(FC_{d})$}}  
            & Fleet Avg Trips & Normalized float [0, 1] (mean trips across all trucks) \\ \cline{2-3}
            & Recent Shovel Usage & Vector of floats (proportion of last 10 decisions per shovel) \\ \cline{2-3}
            & Fleet Diversity & Normalized float [0, 1] (Shannon entropy of decisions) \\ 
        \hline
    \end{tabular}
    \hspace{0.5em}
    \begin{tabular}{|l|c|}
        \hline
        \textbf{Truck Status} & \textbf{Binary Code} \\ 
        \hline
        At Shovel  & 000 \\ 
        At Crusher  & 001 \\ 
        At Dumping Site  & 010 \\ 
        Moving from Shovel to Crusher  & 011 \\ 
        Moving from Shovel to Dumping Site  & 100 \\ 
        Moving from Crusher to Shovel  & 101 \\ 
        Moving from Dumping Site to Shovel  & 110 \\ 
        Breakdown  & 111 \\ 
        \hline
    \end{tabular}
    }
    \caption{\textcolor{black}{Definition of the state-space representation used in the mining fleet simulation. 
Table (a) outlines the structured attributes describing the current system state, including shovel-specific data $(SA_{d})$, active truck attributes $(TA_{d,active})$, and fleet-level contextual variables $(FC_{d})$. 
Table (b) provides the corresponding 3-bit binary encoding scheme for truck operational statuses, used to represent each truck’s current activity or condition within the model.}}
    \label{tab:state_attrib_code}
\end{table}

    \item [2)] \textit{Actions:} Actions are decisions made at each decision step (d), by the RL policy. 

    In our framework, an action corresponds to resource or shovel allocation and the action space is defined as:
    \[
    A_d \in \{1, 2, \dots, SH\},
    \]    
    where \( SH \) represents the total number of shovels, and the action value indicates the shovel ID to which a truck is assigned at decision point \( d \).

    \item [3)] \textit{Rewards:} The reward function quantifies both immediate and cumulative benefits, defining objectives to maximize. In our event-based RL setting, shifts are episodes with sparse rewards given for critical milestones like meeting production targets. The environment features a long-horizon episodic reward and intermediate global rewards to guide learning.

Once the dispatching agent $\mathcal{D}$ outputs action $A_i^d$, the DES environment responds with:

\begin{equation}
R_i^d = r_{\text{imm}}^{d} + \big( r_{\text{epi}}^{d}\; \text{if}\; d = d_{T}\; \text{else}\; 0 \big),
\end{equation}
where $r_{\text{epi}}^{d}$ is the episodic reward at the end of an episode (or shift), and $r_{\text{imm}}^{d}$ is the immediate reward per decision-step $d$.\newline

\textit{Immediate Reward Formulation: }
The immediate reward uses an exponentially weighted sliding window over the latest $k$ decision points:

\textcolor{black}{\subsection{Reward Function Formulation}}

\textcolor{black}{The immediate reward at decision point $d$ incorporates four penalty components:}

\textcolor{black}{
\begin{align}
r^d_{\text{imm}} = -\alpha \cdot \overline{TT}_{\text{Avg}}^{\text{norm}} - \beta \cdot \overline{Q}_{\text{Avg}}^{\text{norm}} - \gamma \cdot (1 - D_{\text{DivScr}}) - \delta \cdot S_{\text{streak}}
\label{eq:reward_imm}
\end{align}
}

\label{eq:reward_imm}
where $\alpha, \beta, \gamma, \delta$ are component weights. Min-max normalization is applied to the exponentially-weighted averages: $x^{\text{norm}} = \frac{x - x_{\text{min}}}{x_{\text{max}} - x_{\text{min}}}$

\textcolor{black}{
\begin{equation}
\overline{TT}_{\text{Avg}} = \sum_{j=1}^{k} \bar{w}_j \cdot \tau_j, \quad \overline{Q}_{\text{Avg}} = \sum_{j=1}^{k} \bar{w}_j \cdot Q^{SH}_j, \quad \bar{w}_j = \frac{e^{\mu \cdot j}}{\sum_{i=1}^{k} e^{\mu \cdot i}}\label{eq:weighted_avg}
\end{equation}
}
where $n = \min(k, |D|)$ is the effective window size with decay parameter $\lambda$. The components are:

\begin{itemize}
    \item \textit{Average Trip Time}: $\tau_d = \frac{1}{|\Omega(d)|} \sum_{\omega \in \Omega(d)} \mathcal{T}_{\omega}(d)$, where $\Omega(d)$ is the set of trucks completing trips and $\mathcal{T}_{\omega}(d)$ is the cycle time for truck $\omega$.
    
    \item \textit{Shovel Queue Time}: $Q_{{SH}_d} = \frac{1}{|\Gamma|} \sum_{\gamma \in \Gamma} \frac{1}{|\Omega_\gamma(d)|} \sum_{\omega \in \Omega_\gamma(d)} Q_{\omega}(d)$, where $\Gamma$ is the set of shovels and $Q_{\omega}(d)$ is the waiting time for truck $\omega$ at shovel $\gamma$.
    
    \item \textit{Diversity Score}: $D_{\text{DivScr}} = H/H_{\max}$, where $H = -\sum_{i=1}^{|\Gamma|} p_i \log_2(p_i + \epsilon)$, $H_{\max} = \log_2(|\Gamma|)$, and $p_i = \text{count}_i/\sum_{j} \text{count}_j$ represents the proportion of assignments to shovel $i$ within a sliding window of size $|W_{\text{div}}|$. This Shannon entropy-based measure quantifies shovel utilization uniformity.
    
    \item \textit{Streak Penalty}: $S_{\text{streak}} = \frac{1}{|W_{\text{streak}}|-1} \sum_{i=1}^{|W_{\text{streak}}|-1} \mathbb{1}(a_i = a_{i-1})$, where $W_{\text{streak}}$ represents the recent $|W_{\text{streak}}|$ decisions, $a_i$ is the assigned shovel at decision $i$, and $\mathbb{1}(\cdot)$ is the indicator function. This penalizes consecutive assignments to the same shovel.
\end{itemize}

\textcolor{black}{The reward function parameters used in the demonstration experiments were determined through preliminary tuning to balance multiple operational objectives: production efficiency, resource utilization uniformity, and queue management. 
Episodic production weight $\omega_{1}=0.4$ and diversity penalty weight $\omega_{2}=0.6$ were applied. A high–performance bonus $b_{1}=0.5$ was triggered when $P_{\text{ratio}}\geq0.95$ and $D_{\text{final}}>0.65$, while a moderate–performance bonus $b_{2}=0.2$ was triggered when $P_{\text{ratio}}\geq0.90$ and $D_{\text{final}}>0.50$. Immediate rewards used a $k=5$ decision–point sliding window with exponential decay $\mu=0.5$.\newline
Complete parameter specifications are documented in Table B.9 (Appendix B). These hyperparameter values are provided as baseline configurations; Mining-Gym users can readily adjust these parameters through configuration files to reflect domain-specific operational priorities or conduct ablation studies on individual reward components.}

The episodic reward at shift termination evaluates overall performance:

\begin{equation}
r_{\text{epi}}^{d_T} = \omega_1 \cdot P_{\text{ratio}} - \omega_2 \cdot (1 - D_{\text{final}}) + B_{\text{perf}} \label{eq:reward_epi}
\end{equation}

where $P_{\text{ratio}} = \min(1, P_{\text{Vol}}/P_{\text{Vol,Target}})$ is the production ratio, $D_{\text{final}}$ is the final diversity score, and the performance bonus is:

\begin{equation}
B_{\text{perf}} = \begin{cases}
b_1 & \text{if } P_{\text{ratio}} \geq \theta_1 \text{ and } D_{\text{final}} > \phi_1 \\
b_2 & \text{if } P_{\text{ratio}} \geq \theta_2 \text{ and } D_{\text{final}} > \phi_2 \\
0 & \text{otherwise}
\end{cases} \label{eq:bonus}
\end{equation}

with bonus values $b_1 > b_2 > 0$ and thresholds $\theta_1 > \theta_2$, $\phi_1 > \phi_2$. The total episode return is:

\begin{equation}
G = \sum_{d=1}^{d_T-1} \gamma^{d-1} r_{\text{imm}}^{d} + \gamma^{d_T-1} \left(r_{\text{imm}}^{d_T} + r_{\text{epi}}^{d_T}\right) \label{eq:total_return}
\end{equation}

where $\gamma \in [0,1]$ is the discount factor, and the RL objective maximizes expected return: $\max_{\theta} \mathbb{E}_{\tau \sim \pi_\theta}[G]$.

\end{itemize}

\section{Implementation Features}\label{Sec:IV}
In this section, we introduce different visualizations and other GUI-based tools that we provide with the Mining-Gym simulator, for better performance comparison and studying change in KPIs.
\subsection{Configuration System}

\begin{table}[h!]
    \centering
    \scalebox{0.7}{
    \begin{tabular}{|l|l|l|}
        \hline
        \multicolumn{2}{|l|}{\textbf{Variables}} & \textbf{Source} \\
        \hline
        \hline
        \multicolumn{3}{|l|}{\textbf{A. Operational Parameters}} \\
        \hline
        \hline
        \multicolumn{2}{|l|}{\textbf{Number of Equipment:}} & - \\
        \hline
        Num. of Trucks & TR & User \\
        \hline
        Num. of Shovels & SH & User \\
        \hline
        Num. of Crushers & CR & User \\
        \hline
        Num. of Dumps & DS & User \\
        \hline
        \multicolumn{2}{|l|}{\textbf{ Queue Size (waiting for resource) $Q$:}} & - \\
        \hline
        At Shovel, Crusher, Dumping site & $Q_{SH}, Q_{CR}, Q_{DS}$ & DES \\
        \hline
        \multicolumn{2}{|l|}{\textbf{Load per trip $L$}} & - \\
        \hline
        Truck carrying Waste, Ore & $L_{W}, L_{O}$ & User \\
        \hline
        \multicolumn{2}{|l|}{\textbf{Truck Speed:}} & - \\
        \hline
        Empty truck, Loaded truck & $S_{\text{EmTR}}, S_{\text{LodTR}}$ & User \\
        \hline
        \multicolumn{2}{|l|}{\textbf{Others:}} & - \\
        \hline
        Number of trips (SH-CR-DS-SH) & $N$ & DES \\
        \hline
        Available (operation unit) time & $T_{\text{SHF}}$ & User \\
        \hline
        Dumping and maneuver time (min) & $T_{\text{DM}}$ & User \\
        \hline
        Shift duration in minutes & $S_{\text{dur}}$ & User \\
        \hline
        Num\_shifts & SN & User \\
        \hline
        \hline
        \multicolumn{3}{|l|}{\textbf{B. Cost and Financial Parameters}}\\
        \hline
        \hline
        Known cost, Estimated cost & $C_{\text{KW}}, C_{\text{EST}}$ & User \\
        \hline
        \hline
        \multicolumn{3}{|l|}{\textbf{C. Equipment Performance and Efficiency}} \\
        \hline
        \hline
        \multicolumn{2}{|l|}{\textbf{Unit Fuel Consumption ($F_{\text{Unit}}$):}} & - \\
        \hline
        Truck with Waste, Ore, Empty & $F_{W}, F_{O}, F_{E}$ & User \\
        \hline
        \multicolumn{2}{|l|}{\textbf{Equipment off time $OF_E$}} & - \\
        \hline
        Truck, Shovel, Crusher, Dumping Site & $OF_{\text{TR}}, OF_{\text{SH}},$ & DES \\
        & $OF_{\text{CR}}, OF_{\text{DS}}$ & \\
        \hline
        \multicolumn{2}{|l|}{\textbf{Equipment idle time $IDL_E$}} & - \\
        \hline
        Truck, Shovel, Crusher, Dumping Site & $IDL_{\text{TR}}, IDL_{\text{SH}},$ & DES \\
        & $IDL_{\text{CR}}, IDL_{\text{DS}}$ & \\
        \hline
        \multicolumn{2}{|l|}{\textbf{Equipment mean time between failure:}} & - \\
        \hline
        Shovel, Truck, Crusher, Dumping Site & $F_{\text{SH}}, F_{\text{TR}}, F_{\text{CR}}, F_{\text{DS}}$ & User \\
        \hline
        \multicolumn{2}{|l|}{\textbf{Equipment mean time to repair:}} & - \\
        \hline
        Shovel, Truck, Crusher, Dumping Site & $R_{\text{SH}}, R_{\text{TR}}, R_{\text{CR}}, R_{\text{DS}}$ & User \\
        \hline
        \hline
        \multicolumn{3}{|l|}{\textbf{D. Loading and Other Time:}} \\
        \hline
        \hline
        Truck loading & TRL & User \\
        \hline
        Dump to Shovel & DTS & User \\
        \hline
        Shovel to Dump & STD & User \\
        \hline
        Truck dumping @ Dump & TRDM & User \\
        \hline
        Crusher to Shovel & CTS & User \\
        \hline
        Shovel to Crusher & STC & User \\
        \hline
        Truck dumping @ Crusher & TRCR & User \\
        \hline
        \hline
        \multicolumn{3}{|l|}{\textbf{E. Other Parameters:}} \\
        \hline
        \hline
        Probability of choosing Crusher & $\epsilon$ & User \\
        \hline
        Target Production volume & $P_{\text{Vol, Targ}}$ & User \\
        \hline
    \end{tabular} }
    \caption{List of variables and parameters used. See Appendix B for detailed parametrization and distributions.}
    \label{tab:variable_sources}
\end{table}

\begin{figure}[htp]
    \centering
    \begin{subfigure}{\linewidth}
        \centering
        \includegraphics[width=0.9\linewidth]{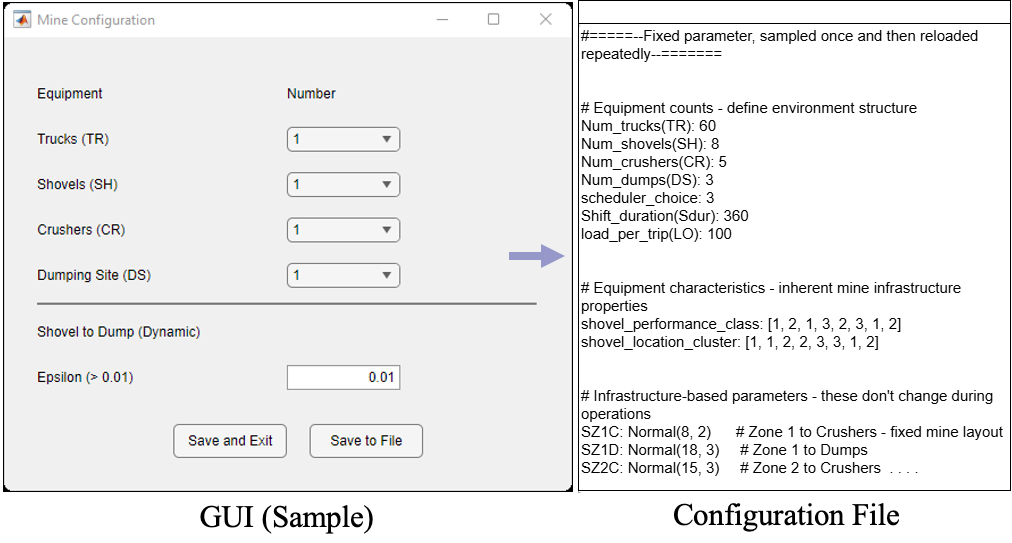}
        \caption{}
        \label{Fig.GUI}
    \end{subfigure}
    
    \vspace{10pt}
    
    \begin{subfigure}{\linewidth}
        \centering
        \includegraphics[width=0.9\linewidth]{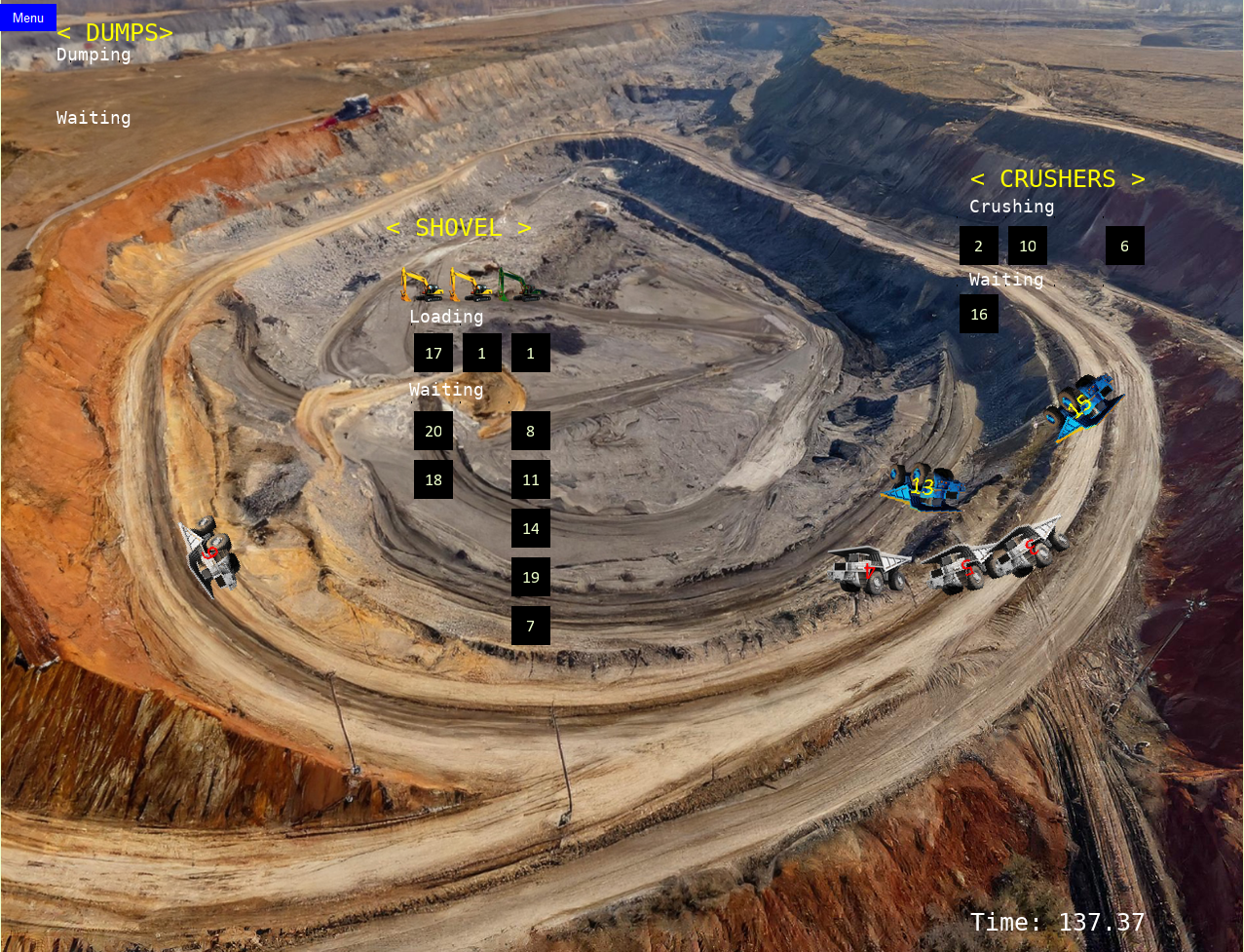}
        \caption{}
        \label{Fig.Visual}
    \end{subfigure}
    
    \caption{(a) Mining-Gym graphical user interface (GUI) and the generated configuration file. (b) Real-time representative visualization of the mine site. The screengrab shows trucks queued at shovels, with the rightmost shovel offline, and trucks transiting between dump sites, crushers, and shovels.}
    \label{Fig.Visual_whole}
\end{figure}

As discussed in the previous section, a large number of parameters must be initialized to run the simulator (see Table~\ref{tab:variable_sources}). To provide an organized method for supplying this information, we have designed a GUI interface. This interface allows users to set values and distributions for stochastic parameters, which are then saved to a human-readable text file. This configuration file is used by the simulator to initialize its parameters during operation (see Fig.~\ref{Fig.GUI}). \textcolor{black}{This file now segregates parameters into fixed infrastructural properties (sampled once per experiment, e.g., number of shovels ) and operational variabilities (stochastically sampled per shift, e.g., fuel consumption and MTBF ). Furthermore, to ensure standardized and reproducible testing across diverse challenges, failure conditions are managed through a dedicated scenario configuration file. This file defines a set of pre-defined stress tests (e.g., loss of one shovel and fifteen trucks) where specific failure parameters are set precisely, allowing for systematic and comparable evaluation of scheduler robustness.}

\subsection{Other Visualization Features}\label{KPI}
We provide a \textit{real-time representative visualization} of the mining setup as configured in the setup file. It features animated truck movements between loading points, crushers, and dumping sites, along with real-time queue formation. The visualization is runnable both during training and when executing a trained model or any other algorithm (see Fig.~\ref{Fig.Visual}).\newline

\textit{Key Performance Indicators (KPIs)}: 
We measure mining efficiency through these essential metrics:

\begin{enumerate}
    \item \textbf{Total production $(\text{shift}^{-1})$:} $P_{Vol} = N \times LO$, where $N$ is the total trips completed and $LO$ is ore load per trip (tons).

    \item \textbf{Trips $(\text{hour}^{-1})$ :} $
    \mathcal{T}_{h} = \frac{60}{\Delta t} \times [N(t) - N(t-\Delta t)] $, 
    where $N(t)$ = cumulative trips at time $t$, $N(t-\Delta t)$ = trips at time $t-\Delta t$, $\Delta t$ = measurement interval (minutes), $h$ is the hour index and $N(t) = \sum_{i=1}^{TR} C_i(t)$ with $TR$ = total trucks, $C_i(t)$ = trips by truck $i$

    \item \textbf{Max. queue length $(\text{hour}^{-1})$
    :} $\widehat{Q}_{SH,h} = \max_{k=1}^{K}\left(\max_{i=1}^{N_{SH}} Q_{SH,i}^{(k)}\right)$, where $h$ is a given hour, $N_{SH}$ the number of shovels, $\mathbf{Q}_{SH}^{(k)}$ the shovel-queue vector at time instant $k$ with element $Q_{SH,i}^{(k)}$ for shovel $i$ ($i=1,\dots,N_{SH}$), and $K$ the number of time instants in hour $h$. Measures bottleneck severity: a high value indicates major congestion events (i.e., long queues) occurred, which points to a potential failure point or poor real-time dispatching.

    \item \textbf{Avg. queue length $(\text{hour}^{-1})$ :} 
    $\overline{Q}_{SH,h}=\frac{1}{K}\sum_{k=1}^{K}\left(\frac{1}{N_{SH}}\sum_{i=1}^{N_{SH}}Q_{SH,i}^{(k)}\right)$.
    Let $h$ be a given hour, $N_{SH}$ the number of shovels, $\mathbf{Q}_{SH}^{(k)}$ the shovel-queue vector at time instance $k$ with element $Q_{SH,i}^{(k)}$ for shovel $i$ ($i=1,\dots,N_{SH}$), and $K$ the number of time-instance in hour $h$. This metric measures system utilization: a low value suggests trucks and shovels are well-utilized, while a high value suggests overall system congestion.

    \item \textbf{Cost per ton $(\text{shift}^{-1})$ :} $\text{CPT} = \dfrac{C_{KW} + C_{EST}}{P_{Vol}}$ \newline
    where $C_{KW}$ = known costs, $C_{EST}$ = estimated costs

    \item \textbf{Fuel consumption :} $FC = \frac{N \times F_{\text{Unit}}}{P_{\text{Vol}}}$ \newline
    where $F_{\text{Unit}} = F_W + F_O + F_E$ with $F_W$, $F_O$, $F_E$ = fuel for waste, ore, empty trips respectively
\end{enumerate}

\begin{figure}[h!]
    \centering
    \includegraphics[width=\linewidth]{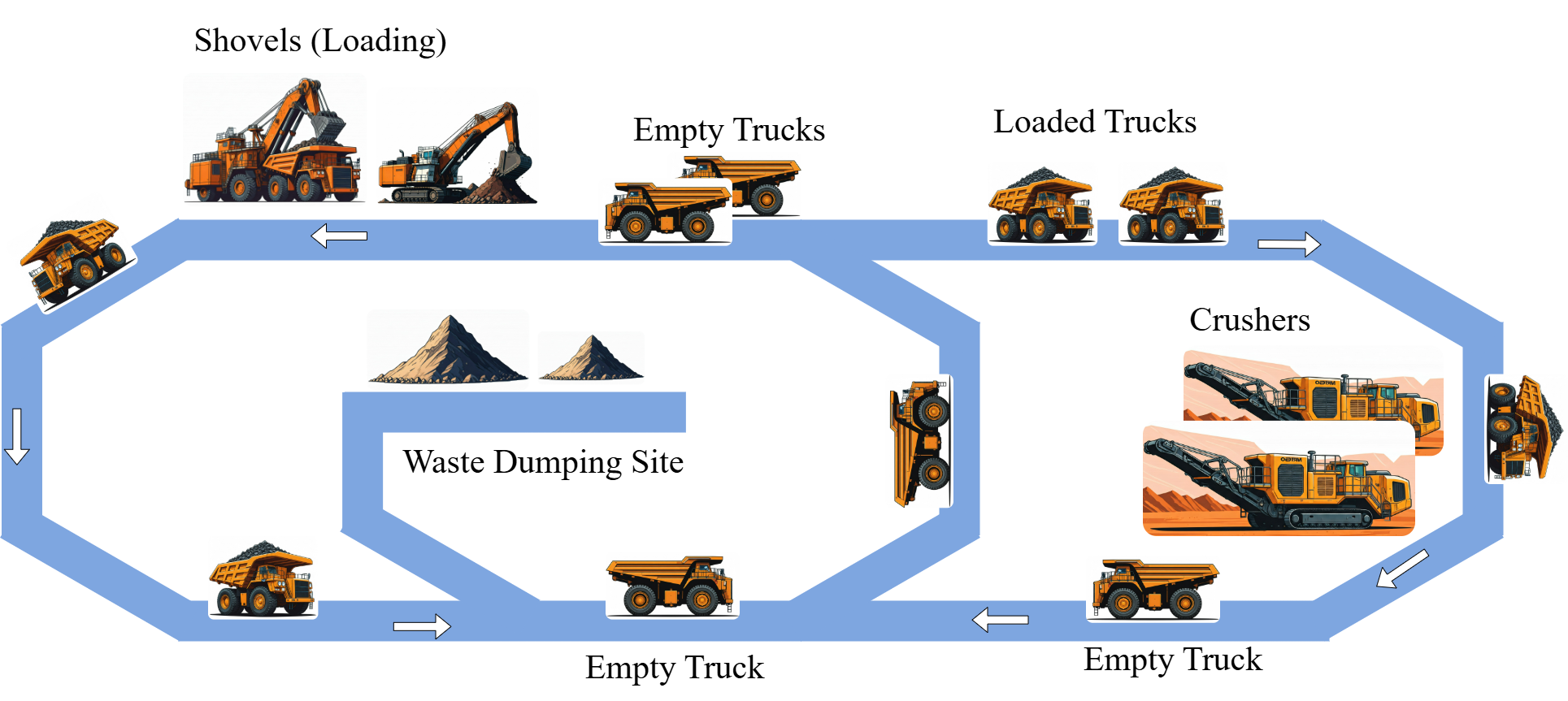}
    \caption{Conceptual Diagram of Minesite}
    \label{fig:minesite_diag}
\end{figure}
    
\section{Experimental Setup} \label{Sec:V}
\subsubsection{Environment Design}
\textcolor{black}{To address scalability and computational efficiency, we configure a large-scale operational scenario with 8 shovels, 60 trucks, 5 crushers, and 3 dumping sites. This represents an upper–medium to large bauxite operation (approximately 5–10 million tonnes annual capacity). The resulting truck–to–shovel ratio of 7.5:1 is within typical industry practice (6–12 trucks per shovel), generating sufficient event density for stress-testing.\newline
With 60 trucks completing 4–6 trips per shift at 100~t payloads, this setup supports approximately 24{,}000–36{,}000~t of daily material movement across three shifts. More importantly for benchmarking, this scale produces high event density: concurrent truck movements, simultaneous loading/dumping, and overlapping breakdown–repair cycles stress-test both the DES engine and RL decision-making under realistic operational loads.\newline
This configuration also enables controlled scalability studies across: (i) computational performance under dense event streams, (ii) algorithmic robustness under resource degradation (e.g., losing 15–30 trucks or 2–3 shovels), and (iii) queue-management complexity under realistic contention. The scale selected represents an upper bound where RL training remains tractable (\= 500 episodes within reasonable runtimes) while still capturing the operational characteristics and challenges of industrial-scale mining logistics.\newline
Equipment specifications are listed in Table~\ref{Tab:Equip_list}, and a conceptual diagram of the minesite is presented in Fig.~\ref{fig:minesite_diag}.}

\begin{table}[ht!]
\centering
\scalebox{0.8}{ 
\begin{tabular}{|l|l|l|l|}
\hline
\hline
\textbf{Equipment} & \textbf{Model/ Make} & \textbf{Quantity} & \textbf{Specification (each)} \\
\hline
\hline
Shovels & CAT 6040 & 8 & 43.7 t bucket size \\
\hline
Trucks & CAT 777 & 60 & 98.2 nominal payload \\
\hline
Crushers & Metso Lokotrack (LT106) & 5 & 450 t / hr capacity \\
\hline
Dumping site & - & 2 & Unlimited \\
\hline
\end{tabular}
}
\caption{Equipment List with Specifications}
\label{Tab:Equip_list}
\end{table}

\subsubsection{Training Setup}
To ensure that the trained agent generalizes well across the testing scenarios, we not only incorporate stochastic parameters to randomize breakdown and repair times for mining equipment and trucks but also enhance the reward function by introducing queue buildup metrics (queue length) and diversity score to prevent repeatedly selecting same shovels. 
%
For training an RL policy, we have used the PPO \cite{schulman2017proximal} algorithm from Stable Baselines3 \cite{stable-baselines3}). We have used all the  default parameters and training settings as  in the repository's PPO implementation. Note the \textbf{scenario parameters} which help specify the failure scenarios to be simulated during training or testing. To validate Mining-Gym as a benchmarking platform and highlight the potential of RL-based truck dispatch scheduling, we present comparative results using a PPO policy trained for 500 episodes (shifts). While a proof-of-concept with limited training, these results suffice to demonstrate both the framework’s feasibility and the performance potential of learned scheduling policies versus classical heuristics.

\subsubsection{Testing Setup} We designed scenarios to test the efficacy of the RL trained scheduler against a vanilla scheduler (random scheduler) as baseline, under different challenging scenarios.\newline
For the random scheduler, the dispatcher’s action \( A_d \) is selected randomly from the available set of shovel IDs \( \{1, 2, \dots, SH\} \). The selection follows a uniform distribution, 
\[
P(A_d = a | S_d = s) = \frac{1}{SH} \quad \text{for all } a \in \{1, 2, \dots, SH\},
\]

i.e. any state \( S_d = s \), each action \( A_d = a \) (where \( a \) is a shovel ID) is equally likely to be chosen, with a probability of \( \frac{1}{SH} \).

\textcolor{black}{The fixed scheduler employs a deterministic round–robin allocation strategy. Each truck $i$ is permanently assigned to a specific shovel based on its ID using a modulo operation:
\begin{equation}
A_{d,i} = (i - 1) \bmod SH
\end{equation}
where $i \in \{1,2,\ldots,TR\}$ represents the truck ID, and $SH$ is the total number of shovels.
This creates a static, pre–determined mapping:
\begin{equation}
P(A_{d,i} = a \mid S_d = s) =
\begin{cases}
1, & \text{if } a = (i - 1) \bmod SH \\
0, & \text{otherwise}
\end{cases}
\end{equation}
Unlike the random scheduler, the fixed scheduler’s allocation is independent of the current system state $S_d$ and remains constant throughout operation, distributing trucks evenly across shovels in a cyclic pattern.
}\newline

In contrast, the RL policy \( \pi_{\theta}(a|s) \) is a learned policy, where the action \( A_d = a \) is selected based on the current state \( S_d = s \) and the parameters \( \theta \). The policy is defined as:

\[
\pi_{\theta}(a|s) = P(A_d = a | S_d = s).
\]

This probability is learned through RL algorithm 
where the agent adjusts \( \theta \) based on rewards from its actions, optimizing its decision-making process over time.
We tested the schedulers under different challenging scenarios, which are governed by a dedicated \textbf{scenario configuration file} to ensure consistent parameter selection and repeatability across all comparative trials.

\renewcommand{\arraystretch}{1.2} 

\begin{table*}[h!]
    \centering
    \caption{Testing Scenarios for Scheduler Robustness Evaluation.}
    \label{tab:testing_scenarios}
    \resizebox{\textwidth}{!}{
        \begin{tabular}{|c|c|l|}
            \hline
            \textbf{Scenario} & \textbf{Failures } & \textbf{Operational Objective} \\
            &\textbf{(\#SH, \#TR)} &\\
            \hline
            \textbf{A} & (0, 0) & \textbf{Baseline:} All equipment available (No failures). \\
            \hline
            \textbf{B} & (0, 15) & \textbf{Moderate Truck Stress:} 25\% of the haul fleet is lost. \\
            \hline
            \textbf{C} & (1, 0) & \textbf{Single Shovel Loss:} Tests response to $\sim$12.5\% loading capacity bottleneck. \\
            \hline
            \textbf{D} & (1, 15) & \textbf{Combined Stress:} Simultaneous constraints on both loading and hauling capacity. \\
            \hline
            \textbf{E} & (2, 30) & \textbf{Severe Disruption:} Pushes the system to critical limits (25\% SH, 50\% TR lost). \\
            \hline
            \textbf{F} & (3, 0) & \textbf{Critical Bottleneck:} Tests scheduling under severe $\sim$37.5\% loading capacity loss. \\
            \hline
        \end{tabular}
    }
\end{table*}

\textcolor{black}{These scenarios systematically vary the number and type of resource failures, replacing the manual parameter definitions.
We tested the schedulers under different challenging scenarios, which are governed by a dedicated \textit{scenarios} to ensure consistent parameter selection and repeatability across all comparative trials. These scenarios systematically vary the number and type of resource failures. \newline
All failure scenarios are repeated according to a over ten repeat trials, with the inherent stochasticity in event timings due to probabilistic distributions. These repeats are then considered in the results section for statistical analysis and discussion. 
Complete experimental scenario parameters, including detailed equipment specifications, stochastic parameters, state space descriptions, and technical specifications for trucks, shovels, and crushers can be found in Appendix C.}

\begin{figure*}[h!]
    \centering
\includegraphics[width=1\linewidth]{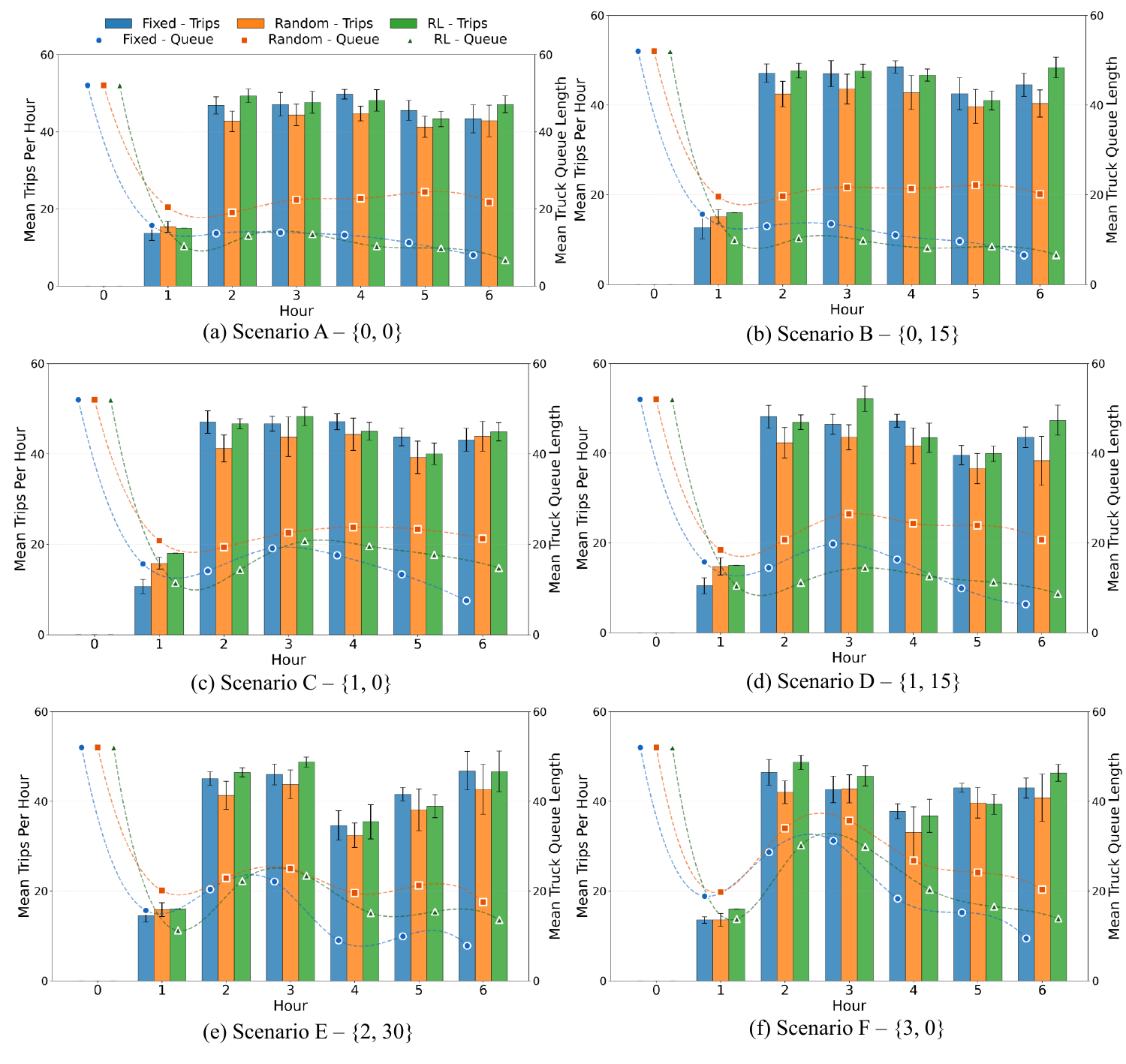}
    \caption{
Comparison of \textit{``Mean Trips per hour''} and \textit{``Mean Truck Queue Length''} over a 6-hour shift under six failure scenarios: 
(a) No failures $\{0,0\}$, 
(b) Moderate Truck Stress $\{0,15\}$, 
(c) Single Shovel Loss $\{1,0\}$, 
(d) Combined Stress $\{1,15\}$, 
(e) Severe Disruption $\{2,30\}$, 
and (f) Critical Bottleneck $\{3,0\}$. 
Shovel failures occurs between 100-150 mins and Truck failures occur between 150-200 mins across the scenarios. Scenarios progress from baseline operation to severe capacity constraints on loading and hauling.
}  \label{fig:trip_metric}
\end{figure*}

\section{Results and Discussion}\label{sec:VI}
\textcolor{black}{Performance across six scenarios is shown in Figure~\ref{fig:trip_metric}, with results averaged over $\mathcal{R}=10$ independent runs. We evaluate schedulers using two metrics, which are averaged versions of two most essential KPIs i.e. hourly max. queue length and trips. The plotted metrics can be defined as follows: \textit{Mean Truck Queue Length} ( $\overline{\widehat{Q}}_{SH,h} ) = \frac{1}{\mathcal{R}} \sum_{r=1}^{\mathcal{R}} \widehat{Q}^{(r)}_{SH,h}$ and \textit{Mean Trips Per Hour} ($\overline{\mathcal{T}}_{h}) = \frac{1}{\mathcal{R}} \sum_{r=1}^{\mathcal{R}} \mathcal{T}^{(r)}_{h}$, where $h$ denotes the hour index and $r$ indexes the run (see  Sec.~\ref{KPI}, for details on the core KPIs and their definitions).
 }
\subsection{Comparing Performance, Efficiency and behavior }
\textcolor{black}{The RL-PPO scheduler demonstrates clear superiority over both fixed and random approaches across all six operational scenarios (A to F), showing the efficacy of RL based dynamic scheduling solution.}\newline

\subsubsection{Productivity Performance :}
\textcolor{black}{The RL scheduler achieves an average improvement of $5.7\%$ in total trips completed compared to the combined performance of fixed and random schedulers. This gain exhibits remarkable consistency across diverse operational conditions, ranging from $4.1\%$ in Scenario C (single shovel loss) to $8.1\%$ in Scenario D (combined shovel loss with moderate truck stress). The most substantial advantages emerge under Combined Stress (Scenario D) and Critical Bottleneck (Scenario F) conditions, demonstrating the RL agent's ability to maximize throughput through superior real-time assignment decisions when resources are constrained.}\newline

\subsubsection{Operational Efficiency :}
\textcolor{black}{The efficiency gains are even more pronounced than productivity improvements. The RL approach reduces the average mean maximum queue length by approximately $24.4\%$ across all scenarios. The magnitude of these reductions varies with operational constraints. The largest improvements occur in scenarios with hauling capacity limitations: Moderate Truck Stress (Scenario B) and Combined Stress (Scenario D), where RL reduces queue lengths by $45.2\%$ and $36.9\%$, respectively. These reductions highlight the RL scheduler's strength in efficiently utilizing limited haul fleet resources and optimally balancing truck distribution. Scenarios with hard loading bottlenecks (C, E, F) exhibit more modest reductions, such as $4.4\%$ in Scenario E, as some queuing becomes unavoidable. Nevertheless, RL minimizes unnecessary waiting and significantly outperforms the random scheduler. These results demonstrate that RL provides a dual benefit: maximizing system output while improving resource efficiency, proving valuable across all operational states from baseline to severe disruptions.}\newline

\subsubsection{Behavioral Characteristics of Scheduling Approaches :}
\textcolor{black}{The behavioral patterns of each scheduling method reveal fundamental differences in their operational dynamics. The RL scheduler's superior performance stems from its immediate and effective response to system disturbances. At system start (Hour 0, initial queue of 52.0 trucks), RL demonstrates rapid stabilization. By Hour 1, RL reduces the average queue to $11.18$, representing a $44.8\%$ improvement compared to Fixed ($16.24$) and Random ($19.85$) schedulers. Despite substantial queue range across the six-hour period (reflecting the severity of Scenarios E and F), the RL scheduler consistently maintains the lowest mean queue length ($14.14$), indicating a stable allocation policy that adapts effectively without excessive oscillation.\newline
The Fixed scheduler shows pronounced instability, evidenced by abrupt oscillations in queue length and a high queue range of $24.89$. Under failure scenarios C–F, simple rule logic cannot compensate for dynamic capacity loss, producing rapid queue spikes followed by sharp overcorrections. This cyclical behavior highlights a structural limitation of static scheduling in non-stationary systems. 
The Random scheduler performs worst overall, with the highest mean queue length ($22.43$) and large queue range ($18.11$), reflecting the absence of coordinated decision logic.coordination.}
\begin{table}[h!]
\centering
\resizebox{\textwidth}{!}{%
\begin{tabular}{|c|c|c|c|c|c|}
\hline
\multirow{2}{*}{\textbf{SID}} 
& \multicolumn{3}{|c|}{\textbf{Performance ($\text{Mean} \pm \text{SD}$)}} 
& \multicolumn{2}{|c|}{\textbf{Statistical Results (Cohen's d)}} \\
\cline{2-6}
& \textbf{RL} & \textbf{Random} & \textbf{Fixed Schedule} 
& \textbf{vs. Random} & \textbf{vs. Fixed} \\
\hline
A & 9000.00 $\pm$ 161.60 & 8111.11 $\pm$ 518.28 & 8877.56 $\pm$ 319.29 & $2.392$ & $0.987$ \\
B & 8622.22 $\pm$ 192.21 & 7900.00 $\pm$ 614.41 & 8555.56 $\pm$ 245.52 & $ 1.587$ & $0.21$ \\
C & 8433.33 $\pm$ 187.08 & 8033.11 $\pm$ 557.77 & 8380.89 $\pm$ 351.58 & $ 1.015$ & $ 0.31$ \\
D & 8566.67 $\pm$ 158.11 & 7622.22 $\pm$ 834.83 & 8377.78 $\pm$ 370.06 & $ 1.572$ & $ 0.664$ \\
E & 7744.44 $\pm$ 339.53 & 7388.89 $\pm$ 527.84 & 7677.78 $\pm$ 164.15 & $ 0.801$ & $ 0.450$ \\
F & 7822.22 $\pm$ 156.35 & 7622.22 $\pm$ 523.87 & 7766.67 $\pm$ 180.28 & $ 0.517$ & $ 0.329$ \\
\hline
\end{tabular}%
}
\caption{Mean $\pm$ SD performance of algorithms. Cohen's d effect size is reported for RL compared with Random and Fixed schedules. Positive d indicates RL superiority; $|d|<0.2$ indicates comparable performance.}
\label{tab:merged_performance_sig_refined}
\end{table}

\subsection{Comparing Throughput}
\textcolor{black}{The RL-PPO scheduler consistently translates these behavioural advantages into superior realized output. Across all six operational settings, RL delivers the highest 6-hour shift production volume, even when capacity is reduced by constrained resources or progressive multi-asset failure. Notably, RL not only outperforms baseline policies in every scenario, but does so with markedly lower output variance — indicating stable exploitation of system capacity rather than brittle opportunism. Under nominal conditions (Scenario A), RL achieves the maximum throughput (9000$\pm$161), validating its efficiency even without disruptions. As failures intensify (Scenarios B–F), RL continues to preserve production margins, with effect sizes (Cohen’s d) remaining positive and typically large — confirming practically meaningful throughput advantage. Even under the most severe disruptions (E, F), where absolute capacity collapse is unavoidable, RL secures the strongest attainable output given the remaining fleet.}

\section{Conclusion and Future Work}\label{sec:VII}
\textcolor{black}{ \textit{Mining-Gym} addresses a key gap in mining process optimization by offering a configurable, open-source benchmarking environment for evaluating reinforcement learning (RL) algorithms in truck dispatch scheduling. It combines high-fidelity discrete-event simulation (DES) modeling, seamless integration with OpenAI Gym, and comprehensive visualization tools. Empirical evaluation across six operational scenarios demonstrates that the RL-PPO scheduler achieves an average improvement of 5.7\% in total trips completed compared to fixed and random scheduling baselines, with gains ranging from 4.1\% under single-asset failure to 8.1\% under combined stress conditions. The RL approach delivers substantial efficiency improvements, reducing mean maximum queue length by approximately 24.4\% on average, with reductions reaching 45.2\% . These productivity and efficiency gains translate to consistently superior throughput performance, with the RL scheduler demonstrating rapid stabilization and adaptive resource allocation even under severe multi-asset disruptions. \textit{Mining-Gym} promotes reproducible research and supports RL adoption in industrial mining applications. \newline
\newline
Future development will focus on several key areas to enhance Mining-Gym's capabilities and research utility. \textbf{Enhanced Fidelity} will incorporate traffic modeling, terrain features, heterogeneous fleets, and operational constraints to improve simulation realism. \textbf{Scenario Expansion} will add test cases that reflect challenges such as weather variability, ore heterogeneity, and dynamic production targets. \textbf{Multi-objective Benchmarks} will enable evaluation of trade-offs among production, cost, equipment longevity, and environmental impact through configurable objective weighting and standardized multi-criteria evaluation protocols. \textbf{Robustness Evaluation Framework} will implement standardized uncertainty quantification protocols including systematic parameter perturbation testing, Monte Carlo robustness analysis, and statistical confidence interval reporting to support algorithmic research in risk-sensitive decision-making. Finally, \textbf{Digital Twin Integration} will connect simulations with real-time operational data for data-driven optimization and practical deployment.}

\section*{Acknowledgments}
This research was partly supported by the Advance Queensland Industry Research Fellowship AQIRF024-2021RD4.

\bibliographystyle{elsarticle-num} 
\bibliography{refer}

\clearpage  
\appendix

\section{Modeling Assumptions and Abstractions}
This section presents the different assumption and abstractions used to simplify the modeling process, in the current version of the simulator
\begin{enumerate}
    \item \textit{Traffic setting and routing:} The current version of the simulation does not incorporate complex vehicle-traffic scenarios. In the future iterations we will aim to integrate this functionality. Currently the simulation assumes a bidirectional and perfectly routed traffic network within the DES modeling, without no traffic congestion.
    \item \textit{Geographical Distribution and Equipment Heterogeneity: } Shovels are distributed across three geographical zones, with 
   zone-specific travel times to crushers and dumps modeled through location cluster assignments. Additionally, 
   shovels have performance classes that determine loading efficiency, affecting cycle times. These characteristics 
   are not observable to the RL agent but influence the underlying  system dynamics (travel times and loading times), creating a 
   partially observable environment where the agent must learn optimal  dispatching without directly knowing individual shovel capabilities 
   or locations.
    \item  \textit{Trailing transport: } Truck cycles transporting tailings to the dumping site are not considered in the current version of the simulator, as they have lower priority and often use separate equipment. Including tailings would add unnecessary complexity without significantly improving the efficiency of primary ore transport operations.
    \item \textit{Environment Reset: } 
    During RL-based training, the environment operates in shifts (episodes). Each shift begins with trucks scheduled to shovels using the Equal-Queue scheduler by default. The shift ends when the time limit is reached, ignoring trucks still in transit. Only completed trips and transported material within the shift's time frame are counted.
\end{enumerate}

\section{Experimental Scenario Parameters}

\begin{table}[htbp]
\centering
\caption{Equipment List with Specifications}
\resizebox{\textwidth}{!}{%
\begin{tabular}{lccl}
\toprule
\textbf{Equipment} & \textbf{Model/Make} & \textbf{Quantity} & \textbf{Specification (each)} \\
\midrule
Shovels per pit & CAT 6040 & 2 (total 4) & 43.7 t bucket size \\
Trucks & CAT 777 & 20 & 98.2 nominal payload \\
Crushers & Metso Lokotrack (LT106) & 3 & 450 t / hr capacity \\
Dumping site & - & 2 & Unlimited \\
\bottomrule
\end{tabular}
}
\end{table}

\begin{table}[htbp]
\centering
\caption{Distributions for stochastic parameters. Parameters are grouped as \textit{Sample Once} (fixed per environment) and \textit{Sample per Episode/Shift} (varies daily). MTBF: Mean Time Between failures, MTTR: Mean Time To Repair.}
\resizebox{\textwidth}{!}{%
\begin{tabular}{lll}
\toprule
\textbf{Parameter Group} & \textbf{Parameter} & \textbf{Distribution / Value} \\
\midrule
\multicolumn{3}{l}{\textbf{A. Sample Once (Environment-fixed)}} \\
\midrule
\multirow{4}{*}{Equipment counts} & Trucks (TR) & 60 \\
 & Shovels (SH) & 8 \\
 & Crushers (CR) & 5 \\
 & Dumps (DS) & 3 \\
\midrule
\multirow{4}{*}{Equipment characteristics} & Truck payload (LO) & 100 t \\
 & Empty Truck Speed (TE) & Normal(55, 12) km/hr \\
 & Loaded Truck Speed (TL) & Normal(30, 8) km/hr \\
 & Shovel bucket capacity (SCP) & Not specified \\
\midrule
\multirow{8}{*}{Equipment Maintenance} & MTBF Shovel (FSH) & Poisson(75) \\
 & MTBF Truck (FTR) & Poisson(90) \\
 & MTBF Crusher (FCR) & Poisson(120) \\
 & MTBF Dumping Site (FDS) & Poisson(180) \\
 & MTTR Shovel (RSH) & Poisson(60) \\
 & MTTR Truck (RTR) & Poisson(45) \\
 & MTTR Crusher (RCR) & Poisson(90) \\
 & MTTR Dumping Site (RDS) & Poisson(30) \\
\midrule
\multirow{2}{*}{Infrastructure travel times} 
 & Shovel → Dump (SZxD) & Zone 1: Normal(18, 3) \\
 &  & Zone 2: Normal(12, 2), Zone 3: Normal(8, 1) \\
 & Shovel → Crusher (SZxC) & Zone 1: Normal(8, 2) \\
 &  & Zone 2: Normal(15, 3), Zone 3: Normal(22, 4) \\
\midrule
\multicolumn{3}{l}{\textbf{B. Sample per Episode / Shift (Varies Daily)}} \\
\midrule
\multirow{3}{*}{Fuel consumption} & Truck with ore (FO) & Normal(0.28, 0.05) lt/ton-km \\
 & Truck empty (FE) & Normal(0.20, 0.06) lt/ton-km \\
 & Shovel (epsilon) & Normal(0.35, 0.07) lt/hr \\
\midrule
\multirow{3}{*}{Truck Loading Times} & Zone 1 (TRL\_C1) & Normal(6, 1) min \\
 & Zone 2 (TRL\_C2) & Normal(9, 2) min \\
 & Zone 3 (TRL\_C3) & Normal(12, 2) min \\
\midrule
\multirow{2}{*}{Truck Dumping / Unloading} & At Dump (TRDM) & Normal(5, 2) min \\
 & At Crusher (TRCR) & Normal(15, 1) min \\
\midrule
\multirow{4}{*}{Experimental Variations} & Shovels to fail (STF) & Uniform(1, 5) \\
 & Trucks to fail (TTF) & Uniform(10, 30) \\
 & Shovel initial breakdown (SIB) & Uniform(10, 45) \\
 & Truck initial breakdown (TIB) & Uniform(10, 90) \\
\bottomrule
\end{tabular}
} \label{tab:Distribution}
\end{table}

\begin{table}[h]
\centering
\caption{Reward function hyperparameters used in demonstration experiments, balancing production efficiency ($\omega_1$, $\omega_2$), performance bonuses and thresholds ($b_1$, $b_2$, $\theta_1$, $\theta_2$, $\phi_1$, $\phi_2$), and immediate reward parameters ($\alpha$, $\beta$, $\gamma$, $\delta$, $k$, $\lambda$, $|W_{\text{div}}|$, $|W_{\text{streak}}|$). All parameters are configurable in Mining-Gym for alternative objectives and ablation studies.}
\label{tab:reward_hyperparams}
\resizebox{\textwidth}{!}{
\begin{tabular}{p{0.40\textwidth} p{0.47\textwidth} c}
\toprule
\textbf{Function Component} & \textbf{Parameter Description} & \textbf{Value} \\
\midrule
\multirow{2}{*}{\parbox[t]{0.40\textwidth}{Episodic Reward (Eq. 5)}} 
& $\omega_1$: Production efficiency weight & 0.4 \\
& $\omega_2$: Diversity penalty weight & 0.6 \\
\midrule
\multirow{4}{*}{\parbox[t]{0.40\textwidth}{Immediate Reward Weights (Eq. 3)}}
& $\alpha$: Trip time penalty weight & 0.25 \\
& $\beta$: Shovel queue time penalty weight & 0.15 \\
& $\gamma$: Diversity score penalty weight & 0.30 \\
& $\delta$: Streak penalty weight & 0.30 \\
\midrule
\multirow{2}{*}{\parbox[t]{0.40\textwidth}{Performance Bonus (Eq. 6)}}
& $b_1$: High-performance bonus value & 0.5 \\
& $b_2$: Moderate-performance bonus value & 0.2 \\
\midrule
\multirow{2}{*}{\parbox[t]{0.40\textwidth}{Production Thresholds (Eq. 6)}}
& $\theta_1$: High production threshold & 0.95 \\
& $\theta_2$: Moderate production threshold & 0.90 \\
\midrule
\multirow{2}{*}{\parbox[t]{0.40\textwidth}{Diversity Thresholds (Eq. 6)}}
& $\phi_1$: High diversity threshold & 0.65 \\
& $\phi_2$: Moderate diversity threshold & 0.50 \\
\midrule
\multirow{3}{*}{\parbox[t]{0.40\textwidth}{Exponentially Weighted Sliding Window (Eq. 4)}}
& $k$: Window size (decision points) & 5 \\
& $\mu$: Exponential decay rate & 0.5 \\
& $n$: Effective window size & $\min(k, |D|)$ \\
\midrule
\multirow{2}{*}{\parbox[t]{0.40\textwidth}{Diversity Score Calculation}}
& $|W_{\text{div}}|$: Recent decisions window & 30 \\
& $\rho$: Numerical stability constant & $10^{-10}$ \\
\midrule
\multirow{1}{*}{\parbox[t]{0.40\textwidth}{Streak Penalty Calculation}}
& $|W_{\text{streak}}|$: Consecutive decision window & 15 \\
\bottomrule
\end{tabular}%
}
\end{table}

\begin{table*}[htbp]
\centering
\caption{Truck Specifications}
\resizebox{\textwidth}{!}{%
\begin{tabular}{lcccc}
\toprule
\textbf{Maker} & \textbf{Model} & \textbf{Payload (MT)} & \textbf{Speed (loaded km/hr)} & \textbf{Fuel Cons. ($t·km^{-1}$)} \\
\midrule
Komatsu & 830E-2 & 231 & 59-64 & 0.3 - 0.4 liters \\
CAT & 793F & 231 & 55-60 & 0.35 - 0.45 liters \\
Terex & MT4400 AC & 218 & 60-64 & 0.35 - 0.45 liters \\
CAT & 777(05) & 98.2 & 62 - 67.1 & 0.25 – 0.35 liters \\
\bottomrule
\end{tabular}%
}
\end{table*}

\begin{table}[htbp]
\centering
\caption{Shovel and Crusher Specifications}
\resizebox{\textwidth}{!}{%
\begin{tabular}{lcc}
\toprule
\textbf{Model} & \textbf{Fuel per Hour} & \textbf{Running Cost (USD)} \\
\midrule
\multicolumn{3}{c}{\textbf{Shovel Specifications}} \\
CAT 6040/6050 Hydraulic Shovels & 250-300 liters per hour & \$3.50 to \$4.50 /t \\
Komatsu PC8000 & 946 liters per hour & \$4.00 - \$5.00 / t \\
Hitachi EX8000-6 & 1100 liters per hour & \$4.50 - \$5.50 / t \\
\midrule
\multicolumn{3}{c}{\textbf{Crusher Specifications}} \\
Metso Lokotrack LT106 & 1.5 to 2 gallons of diesel per hour & \$4.50 to \$6 per hour \\
\bottomrule
\end{tabular}%
}
\end{table}

\newpage
\section{Event-driven or decision-point-based RL
setting}

\begin{figure}
    \centering
    \includegraphics[width=0.9\linewidth]{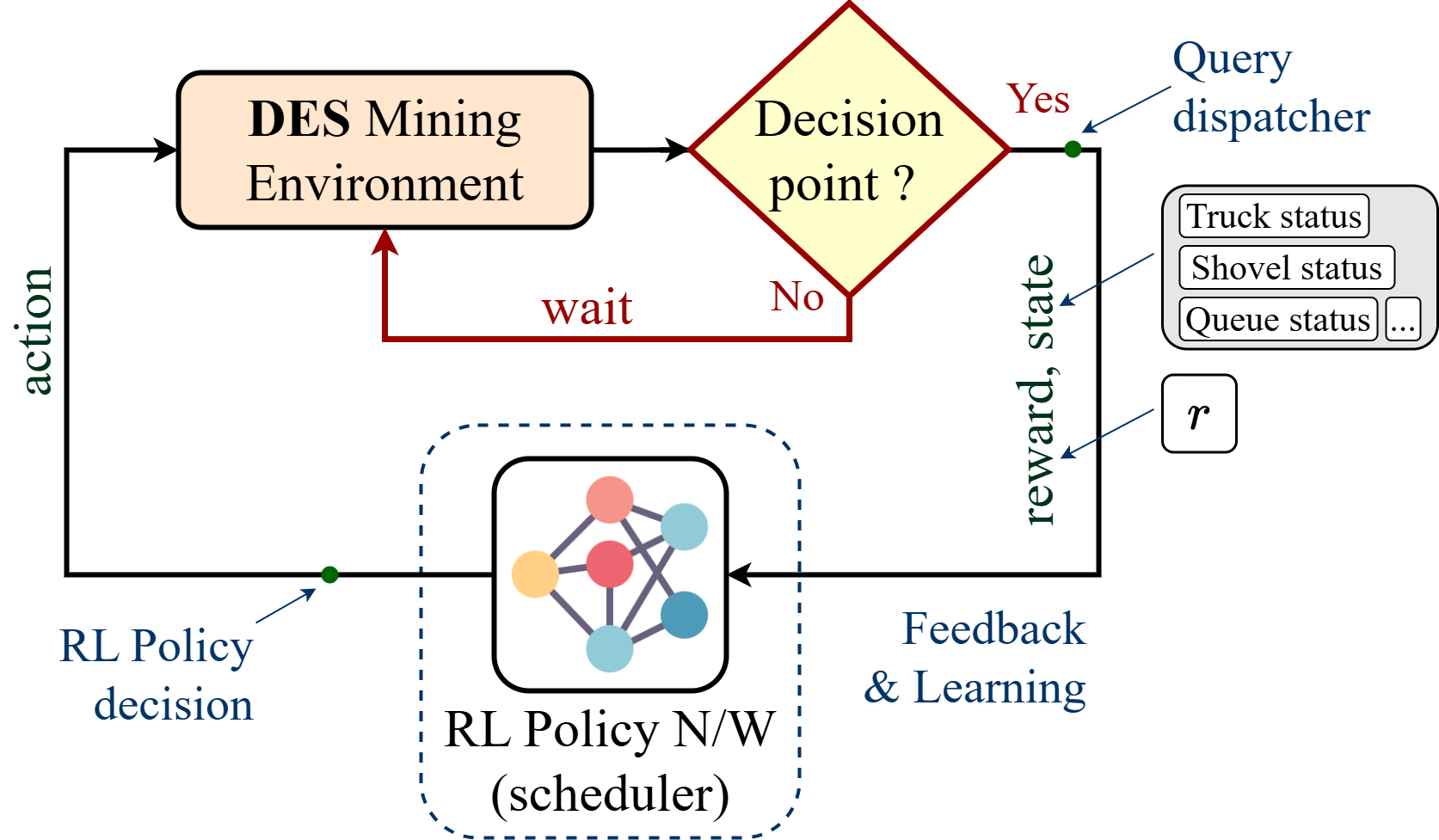}
    \caption{The DES mining environment requests decisions from the RL policy only at specific decision points, such as when
a truck requires a shovel assignment.}
    \label{fig:DES-RL}
\end{figure}

The DES mining environment requests decisions from the RL policy only at specific decision points, such as when a truck requires a shovel assignment. This event-driven approach, illustrated in Fig.\ref{fig:DES-RL}, ensures efficient agent-environment interaction aligned with real mining operations, where continuous interaction is impractical. The agent observes the current state (including shovel and truck attributes) following the Load-Haul-Dump-Return-Query (LHDRQ) cycle, and makes resource allocation decisions using learned dispatching policies. As shown in the figure, the RL policy processes environment states to optimize shovel assignments based on multiple factors, including queue length and equipment status. The simulation handles stochastic elements like equipment failures through the Preemption Handler, while the Resource Handler manages resource availability and assignments. 

\end{document}